\documentclass[10pt,journal,compsoc]{IEEEtran}
\usepackage{cite}
\ifCLASSINFOpdf
   \usepackage[pdftex]{graphicx}
\else
\fi
\usepackage{amsmath}
\usepackage{amssymb}
\usepackage{tabulary,xfrac,enumitem}
\usepackage[dvipsnames]{xcolor}
\usepackage[ruled,noresetcount]{algorithm2e}
\usepackage[pagebackref=false,breaklinks=true,colorlinks,bookmarks=false]{hyperref}
\definecolor{citecolor}{RGB}{34,139,34}

\usepackage{array}
\usepackage{url}

\usepackage{booktabs}
\usepackage{multirow}
\usepackage{threeparttable}
\usepackage{color, colortbl}
\usepackage{changes}
\usepackage{subcaption}
\usepackage{wrapfig}
\usepackage{diagbox}
\usepackage{arydshln} 
\usepackage{makecell}

\makeatletter
\newcommand{\thickhline}{%
    \noalign {\ifnum 0=`}\fi \hrule height 0.5pt
    \futurelet \reserved@a \@xhline
}
\makeatother
\definecolor{LightGray}{gray}{0.9}

\newcommand{\ie}{\textit{i.e.}}
\newcommand{\eg}{\textit{e.g.}}

\newcommand{\vs}{\textit{vs.}\;}
\newcommand{\pub}[1]{{\color{gray}{\tiny{[{#1}]}}}}

\begin{document}
%
\title{Scalable Video Object Segmentation \\ with Identification Mechanism}
%
%
%
%

\newcommand{\zongxin}[1]{{#1}}
\newcommand{\new}[1]{{#1}}

\author{Zongxin~Yang,
        Jiaxu~Miao,
        Yunchao~Wei,
        Wenguan~Wang,
        Xiaohan~Wang,
        and~Yi~Yang
\thanks{
Z. Yang, J. Miao, W. Wang, X. Wang, and Y. Yang are with ReLER, CCAI, Zhejiang University, Hangzhou, China (Email: \{yangzongxin, jiaxumiao, wenguanwang, xiaohan.wang, yangyics\}@zju.edu.cn).
Y. Wei is with the Institute of Information Science, Beijing Jiaotong University, Beijing, China (Email: yunchao.wei@bjtu.edu.cn). Y. Yang is the corresponding author.
}
}

%
%

\markboth{IEEE TRANSACTIONS ON PATTERN ANALYSIS AND MACHINE INTELLIGENCE}%
{Shell \MakeLowercase{\textit{et al.}}: Bare Demo of IEEEtran.cls for Computer Society Journals}
%


\IEEEtitleabstractindextext{%
\begin{abstract}

This paper delves into the challenges of achieving scalable and effective multi-object modeling for semi-supervised Video Object Segmentation (VOS). Previous VOS methods decode features with a single positive object, limiting the learning of multi-object representation as they must match and segment each target separately under multi-object scenarios. Additionally, earlier techniques catered to specific application objectives and lacked the flexibility to fulfill different speed-accuracy requirements. To address these problems, we present two innovative approaches, Associating Objects with Transformers (AOT) and Associating Objects with Scalable Transformers (AOST). 
In pursuing effective multi-object modeling, AOT introduces the IDentification (ID) mechanism to allocate each object a unique identity. This approach enables the network to model the associations among all objects simultaneously, thus facilitating the tracking and segmentation of objects in a single network pass. 
\new{To address the challenge of inflexible deployment, AOST further integrates scalable long short-term transformers that incorporate scalable supervision and layer-wise ID-based attention.
This enables online architecture scalability in VOS for the first time and overcomes ID embeddings' representation limitations.}
Given the absence of a benchmark for VOS involving densely multi-object annotations, we propose a challenging Video Object Segmentation in the Wild (VOSW) benchmark to validate our approaches. We evaluated various AOT and AOST variants using extensive experiments across VOSW and five commonly used VOS benchmarks, including YouTube-VOS 2018 \& 2019 Val, DAVIS-2017 Val \& Test, and DAVIS-2016. Our approaches surpass the state-of-the-art competitors and display exceptional efficiency and scalability consistently across all six benchmarks. Moreover, we notably achieved the $\mathbf{1^{st}}$ position in the 3rd Large-scale Video Object Segmentation Challenge. 
Project page: \url{https://github.com/yoxu515/aot-benchmark}.

\end{abstract}

}

\maketitle
\IEEEdisplaynontitleabstractindextext
%
\IEEEpeerreviewmaketitle

\IEEEraisesectionheading{\section{Introduction}\label{sec:introduction}}

Video Object Segmentation (VOS)~\cite{wang2021survey} plays a crucial role in advancing video comprehension and has great potential in applications such as augmented reality, autonomous vehicles, and video editing. 
It is a challenging task involving extracting, recognizing, and tracking meaningful objects from a continuous video sequence, contributing to a higher-level understanding of dynamic motions and object interactions. 
\new{This paper primarily focuses on semi-supervised/one-shot VOS, a specific problem in which annotated mask(s) for object(s) of interest are provided in the initial frame of the video, and the goal is to track and segment object(s) throughout the entire video sequence.}

Thanks to recent advancements in deep neural networks~\cite{lecun2015deep}, numerous VOS algorithms have been put forward, achieving noteworthy results. Among these methods, STM~\cite{spacetime} and subsequent works~\cite{KMN,EGMN,cheng2021stcn} employ a space-time memory strategy, storing and reading object features of predicted past frames with a memory network. Following the memory reading, they adopt an attention mechanism~\cite{nonlocal} to match the object in the current frame. On the other hand, CFBI(+)~\cite{cfbi,cfbip} and ensuing works~\cite{rpcm,cho2022tackling} make use of global and local matching mechanisms~\cite{feelvos}, which match object pixels or patches from both the first and previous frames and propagate masks to the current frame.

\begin{figure*}[t!]
\begin{center}

\begin{subfigure}[b]{.37\textwidth}
			\centering
			\includegraphics[height=3.3cm]{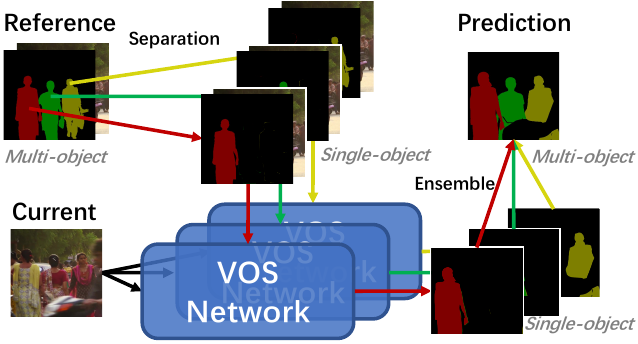}
   \vspace{-1mm}
			\caption{Post-ensemble}\label{fig:post_ensemble}
\end{subfigure}
\hspace{2mm}
\begin{subfigure}[b]{.28\textwidth}
			\centering
			\includegraphics[height=3.4cm]{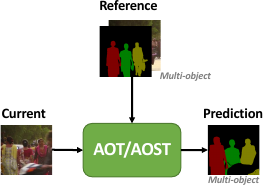}
      \vspace{-1mm}
			\caption{Associating objects (ours)}\label{fig:aot}
\end{subfigure}
\hspace{2mm}
\begin{subfigure}[b]{.22\textwidth}
			\centering
			\includegraphics[height=3.3cm]{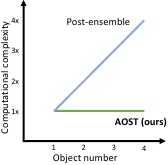}
      \vspace{-1mm}
			\caption{Comparison}\label{fig:complexity}
\end{subfigure}

\end{center}
\vspace{-4mm}
\caption{Post-ensemble methods (\eg,~\cite{spacetime,cfbi}) process each object individually in multi-object scenarios (a). Such single-object modeling is ineffective in exploring multi-object contextual information. In contrast, our approaches associate all objects collaboratively (b), which learns more robust multi-object representation, also resulting in better efficiency (c).} \label{fig:better}
\vspace{-3.5mm}
\end{figure*}

Although the above methods have made significant progress, two important challenges remain: \textbf{1) Single-object Modeling}, \ie, matching and decoding scene features for every single object. Previous methods match each object individually in multi-object scenarios and aggregate all single-object predictions into a multi-object prediction, as illustrated in Fig.~\ref{fig:post_ensemble}. Although such a post-ensemble manner simplifies network designs (since networks are free from accommodating scenes with different object numbers), single-object modeling is ineffective in learning multi-object contextual information. Furthermore, processing numerous objects separately requires significantly more computation than processing a single object. This issue impedes the application of VOS when faced with multi-object scenarios, especially in cases with limited computing resources. \textbf{2) Inflexible Deployment.} Although it is crucial to develop VOS methods that are adaptable and applicable to a wider range of devices and scenarios, \new{previous methods are typically designed for specific objectives, such as improving accuracy~\cite{spacetime, cfbi} or pursuing real-time efficiency~\cite{miles2023mobilevos, realtimevos2, realtimevos1, NEURIPS2020_liangVOS}.} In other words, little attention is paid to the flexibility and scalability of the VOS networks, making it challenging to deploy existing algorithms for devices (such as low-end mobile phones, embedded systems, or high-performance servers) with different speed-accuracy requirements.

To address the above two challenges, a practical approach should possess two essential characteristics: \textbf{1) Multi-object Association}, \ie, collaboratively matching and decoding multiple targets in an end-to-end framework (as illustrated in Fig.~\ref{fig:aot}). By modeling multi-object correlations, matching objects, and decoding multi-object segmentation in a single network pass, the model learns better representations and runs at a single-object-like speed. \textbf{2) Online Scalable Matching and Propagation.} In recent VOS methods (\eg, ~\cite{spacetime,cfbi,cfbip}), the module responsible for matching and propagating object masks is both the key component for achieving better performance and the run-time bottleneck. To realize flexible deployment, we aim to design an online scalable matching and propagation module that allows for straightforward adjustments of speed-accuracy trade-offs and accommodates diverse devices.

In pursuit of multi-object association, we first present \textbf{Associating Objects with Transformers (AOT)}~\cite{aot}, in which the IDentification (ID) mechanism is proposed to assign each object a unique identity and embed all the objects into the same feature space so that the network can learn the associations among all the targets. Moreover, decoding the assigned ID information in the current frame can directly predict multi-object segmentation. To model multi-object associations, a Long Short-Term Transformer (LSTT) is devised to construct hierarchical matching and propagation. Each LSTT block utilizes a long-term attention for matching the reference frame's identities and a short-term attention for matching nearby frames' ones. 

\begin{figure*}[t!]
\centering
\begin{subfigure}[b]{.24\textwidth}
			\centering
			\includegraphics[width=\textwidth]{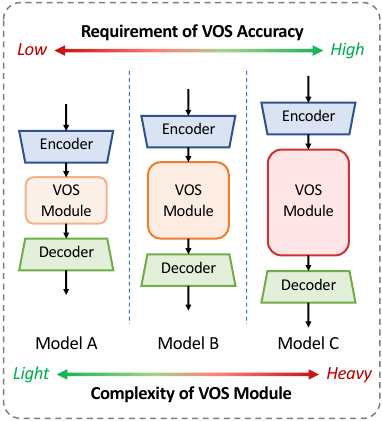}
			\caption{No scalability}\label{fig:before_aost}
\end{subfigure}
\hspace{2mm}
\begin{subfigure}[b]{.24\textwidth}
			\centering
			\includegraphics[width=\textwidth]{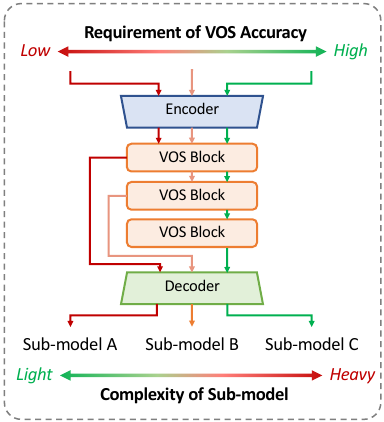}

			\caption{Scalable VOS (ours)}\label{fig:after_aost}
\end{subfigure}
\hspace{2mm}
\begin{subfigure}[b]{.38\textwidth}
			\centering
			\includegraphics[width=\textwidth]{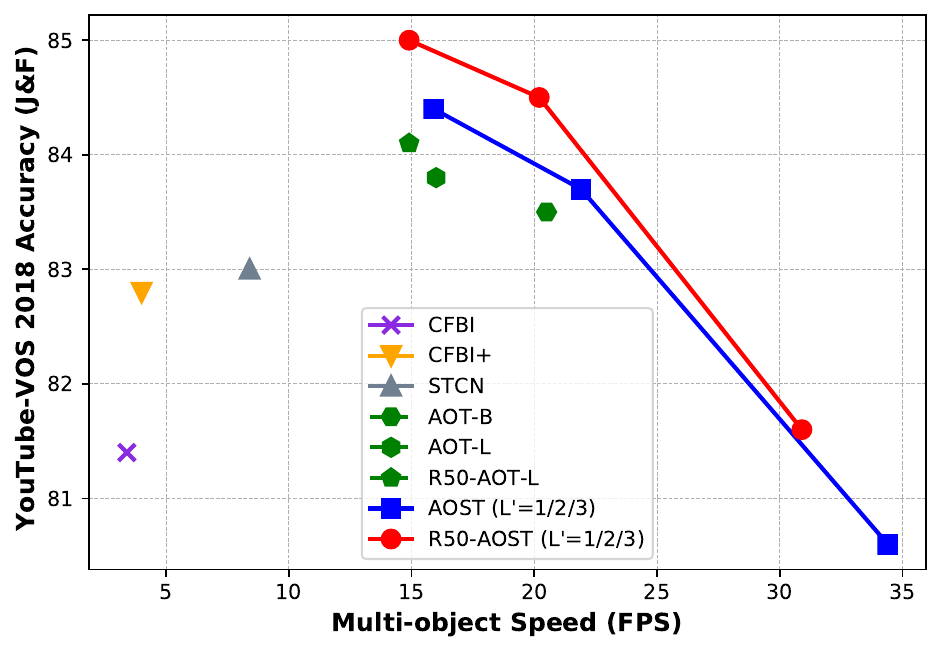}
			\caption{Speed-accuracy Comparisons}\label{fig:speed_accuracy}
\end{subfigure}
\vspace{-2mm}
\caption{(a) Previous VOS methods (\eg,~\cite{cfbip,NEURIPS2020_liangVOS}) are not flexible for different deployment requirements. (b) Our AOST approach supports run-time adjustment for speed-accuracy trade-offs. (c) Under multi-object scenarios, AOST is online scalable and significantly outperforms state-of-the-art competitors~\cite{cfbi,cfbip,cheng2021stcn} on both speed and accuracy. All the results are fairly recorded (using 1 Tesla V100 GPU, full-precision) on YouTube-VOS 2018~\cite{youtubevos}.}\label{fig:aost}
\vspace{-3.5mm}
\end{figure*}

To address the challenge of inflexible deployment, we lift AOT to a more effective and flexible approach, \textbf{Associating Objects with Scalable Transformers (AOST)}, which offers online scalable matching and propagation. 
\new{AOST introduces a Scalable Long Short-Term Transformer (S-LSTT), which is scalable during training and inference. Through scalable supervision, we simultaneously train AOST's sub-networks with different depths efficiently and effectively, enabling online architecture scalability in VOS for the first time. Moreover, AOST incorporates layer-wise ID-based attention to overcome the representation limitation of ID embeddings in multi-layer networks. 
This allows for the online adaptation of the matching and propagation architecture without compromising its performance.}
Fig.~\ref{fig:after_aost} shows that AOST can infer in different transformer depths with test-time trade-offs between state-of-the-art accuracy and efficiency.


Given the absence of a standard VOS benchmark for evaluating algorithms in real-world settings involving dense objects, we present a new \textbf{Video Object Segmentation in the Wild (VOSW)} benchmark to showcase the generalization ability of AOST \& AOT. VOSW is considerably more challenging, incorporating highly dense object annotations and a greater number of object categories. We establish VOSW upon a large-scale panoptic video segmentation dataset~\cite{miao2022large}, employing careful data collection, abundant data clean, and meticulous re-annotation.

To validate the effectiveness and efficiency of our approaches, we conduct extensive experiments on our challenging VOSW and commonly-used benchmarks, including YouTube-VOS 2018 \& 2019 Val~\cite{youtubevos}, DAVIS 2017 Val/Test~\cite{davis2017}, DAVIS-2016~\cite{davis2016}. Benefiting from the multi-object association, AOST \& AOT significantly outperform state-of-the-art competitors on VOSW (\eg, AOST \textbf{75.5\%} \vs STCN 73.1\%). Even using a lightweight encoder~\cite{sandler2018mobilenetv2}, AOST achieves flexible and superior performance (\textbf{80.6$\sim$84.4}\%) on YouTube-VOS 2018 while maintaining multi-object speed $\mathbf{10\times\sim1.8\times}$ faster (\textbf{34.4$\sim$15.9}FPS) than competitors such as CFBI+\cite{cfbip} (82.8\%, 4FPS), as shown in Fig.~\ref{fig:speed_accuracy}.
With a stronger encoder~\cite{swin} and test-time augmentations, we can further boost AOST's performance on commonly-used benchmarks: YouTube-VOS 2018 \& 2019 Val (\textbf{86.5}\%/\textbf{86.5}\%), DAVIS 2017 Val/Test (\textbf{87.0}\%/\textbf{84.7}\%), and DAVIS 2016~\cite{davis2016} (\textbf{93.0}\%). Notably, we ranked $\mathbf{1^{st}}$ in Track 1 (Video Object Segmentation) of the 3rd Large-scale Video Object Segmentation Challenge~\cite{aot_workshop}.

Our contributions in this paper are several folds:
\begin{itemize}[leftmargin=*, noitemsep, topsep=0pt]

\item We embark on a pioneering endeavor to underscore the significance of multi-object modeling and flexible deployment in VOS and propose end-to-end solutions. 

\item To tackle the challenge of single-object modeling, we propose AOT with the identification mechanism to create an effective association of multiple objects. For the first time, multi-object inference in VOS can be efficient as processing a single object.

\item To address the challenge of inflexible deployment, we propose AOST, the first online scalable approach for mask matching and propagation, by incorporating layer-wise ID-based attention and scalable supervision. We can effortlessly scale AOST between state-of-the-art performance and real-time speed during run-time.


\item To validate our identification mechanism, we present VOSW, the first challenging VOS benchmark consisting of real-world scenarios with dense object annotations. 

\item Our approaches demonstrate superior architecture scalability, faster multi-object speed, and remarkable segmentation accuracy on VOSW and commonly-used VOS benchmarks. Notably, we ranked $\mathbf{1^{st}}$ in the 3rd Large-scale Video Object Segmentation Challenge.

\end{itemize}

This paper expands upon our conference paper~\cite{aot}, which has led to multiple follow-up works by ourselves~\cite{zhu2022instance,cheng2023segment,yang2022decoupling,xu2023integrating,yang2024doraemongpt,li2023catr,xu2023video} and other groups~\cite{yu2022batman, xu2023mbptrack, tokmakov2023breaking, mayer2022beyond}. 
Our paper presents extensions in various aspects.
\textbf{(1)} We more precisely state the single-object modeling issue, delve further into the flexible deployment challenge in VOS (\S\ref{sec:introduction}), describe detailed motivations (\S\ref{sec:introduction}, \S\ref{sec:revisit}), summarize the contributions (\S\ref{sec:introduction}), and complete the section of related works (\S\ref{sec:related_works}).
\textbf{(2)} To address the flexible deployment challenge, we introduce the first online scalable approach for object matching and propagation, \ie, AOST (\S\ref{sec:aost}). Through scalable supervision (\S\ref{sec:scalable_supervision}), AOST can infer in different transformer depths to achieve different speed-accuracy trade-offs.
\textbf{(3)} We carefully revisit the LSTT block in \cite{aot} and propose layer-wise identification and gate weights (\S\ref{sec:id_att}) to build AOST with advanced effectiveness. 
\textbf{(4)} We provide more comprehensive descriptions of the algorithms, including more details of the formulations, network architectures (\S\ref{sec:method}), and implementations (\S\ref{sec:implementation}).
\textbf{(5)} To validate our method's multi-object superiority, we introduce a challenging VOS benchmark, VOSW (\S\ref{sec:VOSW}), which covers diverse object categories with dense annotations. 
\textbf{(6)} More extensive experiments are conducted to evaluate AOST variants on VOSW and five commonly-used benchmarks (\S\ref{sec:compare}).
\textbf{(7)} Through additional ablation studies on three aspects (\S\ref{sec:ablation_aost}), we quantitatively demonstrate the effectiveness of the scalable AOST. We also provide more visual results (\S\ref{sec:compare}) for representative success and failure cases.

\vspace{-2.5mm}

\section{Related Work}\label{sec:related_works}

\subsection{Semi-supervised Video Object Segmentation}
In the VOS field~\cite{chen2015video,wang2021survey,liang2023local}, semi-supervised/one-shot VOS algorithms$_{\!}$ track$_{\!}$ objects$_{\!}$ and propagate their first-frame annotations throughout the entire video. Traditional approaches typically involve solving an optimization problem using an energy$_{\!}$ function$_{\!}$ defined$_{\!}$ over$_{\!}$ a$_{\!}$ graph$_{\!}$ structure~\cite{tradition1,tradition3,tradition2}. In recent years, deep learning-based VOS methods utilizing deep neural networks have made significant strides and emerged as the dominant technique in the field.

\noindent\textbf{Finetuning-based Methods.} To narrow the focus of convolutional segmentation networks onto a particular object, finetuning-based methods rely on fine-tuning the networks on the initial frame during test time. Among these methods, OSVOS~\cite{osvos} and MoNet~\cite{xiao2018monet} fine-tune pre-trained networks on the first-frame ground-truth while onAVOS~\cite{onavos} extends the first-frame fine-tuning by introducing an online adaptation mechanism that involves fine-tuning networks with high-confidence predictions during inference online. MaskTrack~\cite{masktrack} uses optical flow to propagate the segmentation mask from one frame to the next while PReMVOS~\cite{premvos} merges four different neural networks using extensive fine-tuning and a merging algorithm. These methods show promising results even though the fine-tuning process seriously slows down all of them during inference.

\noindent\textbf{Template-based Methods.} In order to circumvent the need for test-time fine-tuning, many researchers have treated annotated frames as templates and explored ways to use or match template features to track objects. For instance, OSMN~\cite{osmn} deploys a network that extracts an object embedding and another one that predicts segmentation based on said embedding. PML~\cite{pml} learns a pixel-wise embedding with the nearest neighbor classifier, whereas VideoMatch~\cite{videomatch} uses a matching layer to map pixels of the current frame to those of the annotated frame in a learned embedding space. Similarly, CFBI(+)\cite{cfbi,cfbip,yang2019going,yang2020cfbi+} and following works~\cite{rpcm,cho2022tackling} extend the pixel-level matching mechanism by locally matching with the previous frame~\cite{feelvos} and collaboratively matching background objects. RGMP\cite{rgmp} also harvests guidance information from both the first frame and the previous one, but using a siamese encoder with two shared streams. Additionally, RPCM~\cite{rpcm} proposes a correction module to bolster the reliability of pixel-level matching. Instead of using matching mechanisms, some methods~\cite{LWLVOS,mao2021joint} suggest using an online few-shot learner to identify features of the given object.

\noindent\textbf{Attention-based Methods.} Recent approaches have focused on modeling attention amongst video frames utilizing advanced attention mechanisms~\cite{att,transformer,nonlocal}, which facilitate the propagation of target information from segmented frames to the current frame. To achieve this, STM~\cite{spacetime} and its subsequent works (\eg, \cite{KMN,seong2022video,cheng2021stcn,cheng2022xmem,zhang2023boosting,rde_cvpr22}) employ a memory network to integrate past-frame predictions into memory and implement non-local attention mechanisms to propagate object information from the memory to the current frame. \new{Differently, 
ISVOS~\cite{wang2023look} uses transformers~\cite{transformer} to introduce instance understanding with instance queries,
which are object-agnostic. By contrast, our S-LSTT propagates and aggregates object information in past frames.}

Most VOS methods tend to learn to decode features by considering a single positive object, which leads to the segmentation of each target separately under multi-object scenarios. This approach is inefficient for learning and tracking multiple objects' correlations. Although some methods~\cite{cfbi,cfbip,cheng2021stcn,wang2018semi} relieve this problem by sharing backbone features and pixel-matching maps across objects, these methods still compute mask propagation and segmentation individually for different objects. \new{Apart from the single-object modeling issue, previous methods are generally designed for specific application goals, such as enhancing accuracy~\cite{spacetime,feelvos,cfbi,mao2021joint} or pursuing real-time efficiency\cite{miles2023mobilevos,realtimevos2,realtimevos1,NEURIPS2020_liangVOS}}. Such methods are difficult to flexibly adapt their architectures for real-world deployments with different performance requirements or computational limitations.

\new{Although there have been researchers who developed dynamic and flexible image networks~\cite{teerapittayanon2016branchynet,bolukbasi2017adaptive,yu2018slimmable}, rare studies pay attention to network flexibility and scalability in VOS.
To overcome these challenges, we propose AOST, which simultaneously associates and decodes multiple objects while offering runtime speed-accuracy scalability.}





\begin{figure*}[t!]
\begin{center}

\begin{subfigure}[b]{.285\textwidth}
			\centering
   \vspace{-3mm}
			\includegraphics[width=\textwidth]{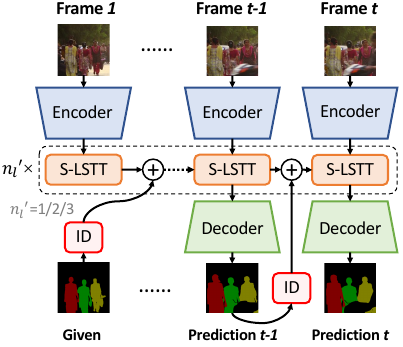}
			\caption{Inference}\label{fig:overview}
\end{subfigure}
\hspace{1mm}
\begin{subfigure}[b]{.197\textwidth}
			\centering
			\includegraphics[width=\textwidth]{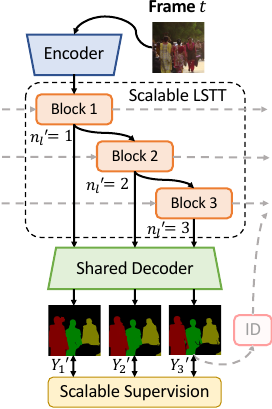}
			\caption{Training}\label{fig:overview_training}
\end{subfigure}
\hspace{1mm}
\begin{subfigure}[b]{.233\textwidth}
			\centering
                \vspace{-1mm}
			\includegraphics[width=\textwidth]{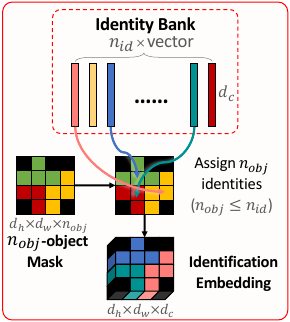}
			\caption{Identity assignment}\label{fig:id}
\end{subfigure}
\begin{subfigure}[b]{.243\textwidth}
			\centering
			\includegraphics[width=\textwidth]{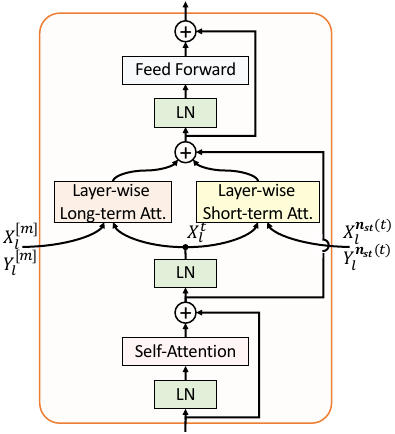}
			\caption{S-LSTT block}\label{fig:lstt}
\end{subfigure}

\end{center}
\vspace{-4mm}
\caption{\textbf{The overview of AOST (\S\ref{sec:method})}. (a)(b) Inference and training of AOST. The multi-object masks are embedded by using our identification mechanism. Moreover, an S-LSTT with dynamic depth ($n_l'=1/2/3$) is responsible for matching multiple objects collaboratively and hierarchically. (c) An illustration of the IDentity assignment (ID) designed for transferring a $n_{obj}$-object mask into an identification embedding. (d) The structure of an S-LSTT block. LN: layer normalization~\cite{ln}.}\label{fig:aot_overview}
\vspace{-3mm}
\end{figure*}

\vspace{-3.0mm}
\subsection{Transformers in Video Applications} 
Transformer$_{\!}$ architecture$_{\!}$ is$_{\!}$ firstly$_{\!}$ proposed$_{\!}$ in~\cite{transformer}$_{\!}$ for constructing hierarchical attention-based networks in machine translation. It employs attention mechanisms to compute correlations with input elements and aggregate their information, enabling global correlation modeling in parallel and leading to better memory efficiency than models that use Recurrent Neural Networks (RNNs).$_{\!}$ Due to their effectiveness in natural language processing$_{\!}$$_{\!}$ (NLP)$_{\!}$ tasks$_{\!}$~\cite{devlin2018bert,radford2019language,synnaeve2019end}$_{\!}$,$_{\!}$ Transformer$_{\!}$ networks$_{\!}$ have$_{\!}$ been$_{\!}$ widely$_{\!}$ used$_{\!}$. Recently,$_{\!}$ Transformer$_{\!}$ blocks$_{\!}$ have$_{\!}$ been introduced in many computer vision tasks such as image classification~\cite{vit,vaswani2021scaling,swin}, object detection$_{\!}$~\cite{detr}/segmentation$_{\!}$~\cite{vistr}$_{\!}$, and image generation~\cite{parmar2018image}, producing promising results compared to Convolutional Neural Network (CNN)-based networks.


The remarkable success of image transformers has sparked a growing interest in exploring attention mechanisms for video-based recognition tasks~\cite{arnab2021vivit,zhu2021temporal,lu2023show,wang2022align}. Certain techniques~\cite{bertasius2021space,arnab2021vivit} have extended image transformers to enable spatial-temporal modeling, demonstrating significant performance on video action recognition. Other methods~\cite{fan2021multiscale,liu2022video} have sought to minimize the redundancy of spatial information by constructing multi-scale vision transformers for video processing. For object-centric video tasks, including visual object tracking, many transformer-based methods~\cite{wang2021transformer,chen2021transformer,yan2023universal,yan2022towards} have been proposed. While several encoder-decoder-based transformers~\cite{meinhardt2022trackformer,chu2023transmot} have shown success in tracking multiple objects by jointly detecting and tracking objects, they are limited by their dependency on pre-defined object categories, such as humans and vehicles, and they predict trajectories in bounding-box format, rather than more precise pixel-level segmentation.

Several$_{\!}$ previous$_{\!}$ VOS$_{\!}$ methods$_{\!}$~\cite{lin2019agss,spacetime,liu2022global,liu2022learning}$_{\!}$ have$_{\!}$ utilized$_{\!}$ attention$_{\!}$ mechanisms$_{\!}$ to$_{\!}$ match$_{\!}$ object$_{\!}$ features$_{\!}$ and$_{\!}$ propagate segmentation masks from past frames to present ones. However,$_{\!}$ these$_{\!}$ methods$_{\!}$ only$_{\!}$ consider$_{\!}$$_{\!}$ one$_{\!}$ positive$_{\!}$ target$_{\!}$ in$_{\!}$ attention$_{\!}$ processes,$_{\!}$ and$_{\!}$ hierarchy-based$_{\!}$ attention$_{\!}$ has$_{\!}$ not$_{\!}$ been$_{\!}$ well$_{\!}$ studied.$_{\!}$ This$_{\!}$ paper$_{\!}$ introduces$_{\!}$ a$_{\!}$ Scalable$_{\!}$ Long$_{\!}$ Short-Term$_{\!}$ Transformer$_{\!}$ (S-LSTT)$_{\!}$ architecture$_{\!}$ that$_{\!}$ can$_{\!}$ efficiently$_{\!}$ construct$_{\!}$ multi-object$_{\!}$ matching$_{\!}$ and$_{\!}$ propagation$_{\!}$ within$_{\!}$ hierarchical$_{\!}$ structures$_{\!}$ for$_{\!}$ VOS,$_{\!}$ achieving$_{\!}$ a$_{\!}$ balance$_{\!}$ between$_{\!}$ segmentation$_{\!}$ accuracy$_{\!}$ and$_{\!}$ efficiency$_{\!}$ during$_{\!}$ testing.

\vspace{-3.5mm}

\section{Revisit Previous VOS Solutions}\label{sec:revisit}

In VOS, many video scenarios have multiple targets or objects required for tracking and segmenting. However, these methods primarily focus on matching and decoding a single object, then ensemble all the single-object predictions into a multi-object prediction, as illustrated in Fig.~\ref{fig:post_ensemble}. Let $\mathrm{\mathbf{F^{vos}}}(\cdot)$ be a VOS network for single-object segmentation prediction, and $\mathrm{\mathbf{Ens}}(\cdot)$ be an ensemble function, such as $\mathrm{\mathbf{Softmax}}$ or soft aggregation\cite{spacetime}. The formula for such post-ensemble manner, which processes $n_{obj}$ objects, is:
\begin{equation*}\small
   \new{\hat{Y}^t=\mathrm{\mathbf{Ens}}(\mathrm{\mathbf{F^{vos}}}(I^t|I^{[m]},Y^{[m]}_1),...,\mathrm{\mathbf{F^{vos}}}(I^t|I^{[m]},Y^{[m]}_{n_{obj}})),}
\end{equation*}
where $I^t$ and $I^{[m]}$ represent the image of the current frame and memory frames (with frame indices $[m]$), respectively. ${Y^{[m]}_1,...,Y^{[m]}_{n_{obj}}}$ are the memory masks containing the given reference mask and past predicted masks of all $n_{obj}$ objects. This approach extends single-object networks into multi-object applications, rendering it unnecessary to adapt the network to different object numbers.

While post-ensemble is straightforward, it has two limitations. \textbf{(i)}$_{\!}$ When$_{\!}$ processing$_{\!}$ multiple$_{\!}$ objects$_{\!}$ separately,$_{\!}$ the post-ensemble$_{\!}$ manner$_{\!}$ requires$_{\!}$ multiple$_{\!}$ times$_{\!}$ computational$_{\!}$ costs for$_{\!}$ processing$_{\!}$ a$_{\!}$ single$_{\!}$ object$_{\!}$.$_{\!}$ This$_{\!}$ issue$_{\!}$ hinders$_{\!}$ the$_{\!}$ learning$_{\!}$ of$_{\!}$ multi-object$_{\!}$ representations$_{\!}$ and$_{\!}$ the$_{\!}$ application$_{\!}$ of$_{\!}$ VOS$_{\!}$ in$_{\!}$ multi-object$_{\!}$ scenarios$_{\!}$.$_{\!}$ \textbf{(ii)}$_{\!}$ The$_{\!}$ architecture$_{\!}$ of$_{\!}$ $\mathrm{\mathbf{F^{vos}}}(\cdot)$$_{\!}$ is$_{\!}$ designed$_{\!}$ for$_{\!}$ specific$_{\!}$ objectives,$_{\!}$ such$_{\!}$ as$_{\!}$ improving$_{\!}$ accuracy$_{\!}$ or$_{\!}$ pursuing$_{\!}$ real-time$_{\!}$ efficiency, and is not flexible to be scaled for different speed-accuracy requirements.


To overcome these challenges, we expect a solution to be capable of associating multiple objects uniformly instead of individually while offering online architectural flexibility. 
As$_{\!}$ depicted$_{\!}$ in$_{\!}$ Fig.~\ref{fig:aot},$_{\!}$ our$_{\!}$ AOST$_{\!}$ can$_{\!}$ associate$_{\!}$ and$_{\!}$ segment$_{\!}$ multiple$_{\!}$ objects$_{\!}$ within$_{\!}$ an$_{\!}$ end-to-end$_{\!}$ framework,$_{\!}$ rendering$_{\!}$ the$_{\!}$ processing$_{\!}$ of$_{\!}$ multiple$_{\!}$ objects$_{\!}$ just$_{\!}$ as$_{\!}$ efficient$_{\!}$ as$_{\!}$ processing$_{\!}$ a$_{\!}$ single$_{\!}$ one$_{\!}$ (Fig.$_{\!}$\ref{fig:complexity}).$_{\!}$ Moreover,$_{\!}$ our$_{\!}$ AOST$_{\!}$ model$_{\!}$ yields$_{\!}$ improved performance under multi-object$_{\!}$ scenarios by uniformly learning contrastive embeddings among multiple objects'$_{\!}$ regions.$_{\!}$ Additionally, our model's$_{\!}$ training is more efficient due to its ability to concurrently learn to infer at different depths while considering the trade-off between speed and accuracy, achieved through scalable supervision using long short-term transformers. The AOST framework can balance the VOS matching and propagation between real-time speed and state-of-the-art accuracy during inference. By contrast, previous methods with no network scalability design cannot achieve this capability (Fig.~\ref{fig:before_aost}).

\vspace{-3.5mm}

\section{Methodology}\label{sec:method}

In this section, we first present our proposed identification mechanism for multi-object modeling in VOS. Subsequently, we introduce the Long Short-Term Transformer (LSTT) designed for forming hierarchical multi-object associations, which are equipped by our Associating Objects with Transformers (AOT) approach. Based on our LSTT, we further introduce Scalable Long Short-Term Transformer (S-LSTT) to pursue the online scalable VOS approach, Associating Objects with Scalable Transformers (AOST). Combined with newly proposed layer-wise ID-based attention and scalable supervision, S-LSTT possesses an architecture with scalable depth and allows for multiple accuracy-efficiency trade-offs during run-time with significant performance improvement over our AOT. An overview of AOST is illustrated in Fig.~\ref{fig:overview}.


\vspace{-3.0mm}
\subsection{Associating Objects with Transformers}\label{sec:aot}

\subsubsection{IDentification (ID) Mechanism}\label{sec:id_mec}

Several$_{\!}$ recent$_{\!}$ methods$_{\!}$ have$_{\!}$ utilized$_{\!}$ attention$_{\!}$ mechanisms$_{\!}$ and$_{\!}$ shown$_{\!}$ promising$_{\!}$ results$_{\!}$~\cite{spacetime,EGMN,KMN}.$_{\!}$ To$_{\!}$ define$_{\!}$ these$_{\!}$ methods,$_{\!}$ we$_{\!}$ introduce$_{\!}$ $Q \in \mathbb{R}^{d_h d_w \times d_c}$,$_{\!}$ $K \in \mathbb{R}^{d_t d_h d_w \times d_c}$,$_{\!}$ and$_{\!}$ $V \in \mathbb{R}^{d_t d_h d_w \times d_c}$$_{\!}$ as the query of the current frame, the key of the memory frames, and the value of the memory frames, respectively, where $d_t$, $d_h$, $d_w$, and $d_c$ refer to the temporal, height, width, and channel dimensions. An attention-based matching and propagation can be formulated as:
\vspace{-1.0mm}
\begin{equation}\small
\label{equ:att}
    \mathrm{\mathbf{Att}}(Q,K,V)=\mathrm{\mathbf{Corr}}(Q,K)V=\mathrm{\mathbf{Softmax}}(\frac{QK^{tr}}{\sqrt{C}})V,
\end{equation}
where a matching map is first calculated by the correlation function $\mathrm{\mathbf{Corr}}(\cdot,\cdot)$, and the value embedding $V$ is then propagated into each position of the current frame.



In such single-object propagation, the binary mask information stored in memory frames is integrated into $V$ using an additional memory encoder network. This enables the information to be transmitted to the current frame by applying Eq.~\ref{equ:att}. Once the value feature $V$ has been propagated, a convolutional decoder network predicts the probability logit of the current frame's single-object segmentation.

To realize more effective and efficient multi-object modeling, the primary challenge of propagating and decoding multi-object information in an end-to-end network is how to modify the network to handle various targeted numbers. To tackle this issue, we suggest an identification mechanism that comprises identification embedding and decoding utilizing attention mechanisms.


\noindent\textbf{Identification Embedding} refers to the approach of embedding the masks of various objects in a single feature space to promote multi-object propagation. As depicted in Fig.~\ref{fig:id}, the method begins with the creation of an identity bank, denoted by the symbol $D \in \mathbb{R}^{n_{id} \times d_c}$, consisting of $n_{id}$ identification vectors with $d_c$ dimensions. Each object is randomly assigned a distinct identification vector from $D$ to embed different object masks. Suppose a video sequence has $n_{obj}$ ($n_{obj}<n_{id}$) objects in the scene. In that case, the one-hot mask of all the $n_{obj}$ objects, $Y \in \{0,1\}^{d_t d_h d_w \times n_{obj}}$, is embedded into an identification embedding, $E \in \mathbb{R}^{d_t d_h d_w \times d_c}$, by randomly assign each object region with a distinct identification vector selected from the bank $D$. The identification embedding $E$ of $n_{obj}$ objects' masks can be formulated as:
\vspace{-1.0mm}
\begin{equation}\small
\label{equ:id}
\vspace{-1.0mm}
    E=\mathrm{\mathbf{ID}}(Y,D)=YPD,
\vspace{-1.0mm}
\end{equation}
where $P \in \{0,1\}^{n_{obj} \times n_{id}}$ represents a random permutation matrix that ensures that $P^{tr}P$ is a $n_{id} \times n_{id}$ unit matrix, allowing for the random selection of $n_{obj}$ distinct identification vectors. The application of $\mathrm{\mathbf{ID}}(\cdot,\cdot)$ assignment ensures that different objects have unique identification embeddings. Then, we can propagate multi-object identification information from memory frames to the current frame by attaching the identification embedding $E$ with the value feature $V$ in the attention. One approach to attaching $E$ directly to $V$ is to add it to $V$ through the equation below:
\begin{equation}\small
\label{eq: attid}
\vspace{-1.0mm}
\begin{aligned}
    V'&=\mathrm{\mathbf{Att}}(Q,K,V+\mathrm{\mathbf{ID}}(Y,D)) \\
    &=\mathrm{\mathbf{Att}}(Q,K,V+E),
\end{aligned}
\vspace{-0.5mm}
\end{equation}
where $V' \in \mathbb{R}^{d_h d_w \times d_c}$ aggregates all the visual and identification embeddings of $n_{obj}$ objects from the matching and propagation process.



\noindent\textbf{Identification Decoding} follows closely and is designed to predict the probabilities of all the objects based on the aggregated feature $V'$. We firstly employ a convolutional decoding network $\mathrm{\mathbf{F^{dec}}}(\cdot)$ to predict the probability logit for all the $n_{id}$ identities in the bank $D$. We then select the assigned $n_{obj}$ identities and calculate the probabilities for $n_{obj}$ objects. This process is represented as follows:
\begin{equation}\small
    Y'=\mathrm{\mathbf{Softmax}}(P\mathrm{\mathbf{F^{dec}}}(V'))=\mathrm{\mathbf{Softmax}}(PL),
\end{equation}
where $L \in \mathbb{R}^{d_h d_w \times n_{id}}$ represents the probability logits for all $n_{id}$ identities, $P$ is the same permutation matrix used in the identity assignment (Eq.~\ref{equ:id}), and $Y' \in \{0,1\}^{d_h d_w \times n_{obj}}$ represents the probability prediction for all $n_{obj}$ objects.

\begin{figure}[t!]
\begin{center}

\begin{subfigure}[b]{.45\columnwidth}
			\centering
			\includegraphics[width=0.9\columnwidth]{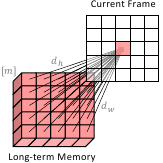}
   \vspace{-1mm}
			\caption{Long-term Attention}\label{fig:long_term}
\end{subfigure}
\begin{subfigure}[b]{.45\columnwidth}
			\centering
			\includegraphics[width=0.9\columnwidth]{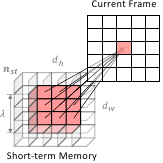}
   \vspace{-1mm}
			\caption{Short-term Attention}\label{fig:short_term}
\end{subfigure}

\end{center}
\vspace{-4mm}
\caption{\textbf{Illustrations of long-term attention and short-term attention (\S\ref{sec:LSTT})}. (a) Long-term attention employs a non-local manner to match all the locations in long-term memory. (b) In contrast, the short-term attention only focuses on a nearby spatial-temporal region with a shape of ${n_{\mathrm{ST}}}\lambda^2$.}\label{fig:long_short_term}
\vspace{-3.0mm}
\end{figure}

\vspace{-2.0mm}
\subsubsection{Long Short-Term Transformer (LSTT)}\label{sec:LSTT}

\new{Previous methods~\cite{spacetime,KMN} always utilize only one layer of attention (Eq.~\ref{equ:att}) to aggregate information about a single object. However, these methods are not ideal and cannot dynamically scale for different deployment requirements. Thus, we carefully design the Long Short-Term Transformer to realize hierarchical object matching and propagation.}

\new{Following the common transformer blocks~\cite{transformer,devlin2018bert}, the LSTT block starts from a self-attention layer. Then, we additionally introduce a long-term attention, for aggregating targets' information from long-term memory frames and a short-term attention, for learning temporal smoothness from nearby short-term frames. The final module is based on a common 2-layer feed-forward MLP with GELU~\cite{gelu} non-linearity in between. All the attention modules are implemented in multi-head attention~\cite{transformer}. Nevertheless, we only introduce their single-head formulas below for simplicity.}

\noindent\textbf{Long-Term Attention} is in charge of collecting information about objects from memory frames, which include the reference frame and stored predicted frames, and propagating it to the current frame. Due to the varying time intervals between the current and past long-term frames, temporal smoothness is challenging to ensure. As a result, long-term attention utilizes a non-local attention, \ie, Eq.~\ref{equ:id}. 

Let $X_l^{t} \in \mathbb{R}^{d_h d_w \times d_c}$ denote the input feature for time $t$ and $l$-th LSTT layer, where $l \in {1,...,n_l}$ is the index of LSTT blocks. The formula for long-term attention is as follows: 
\begin{equation}\small\label{eq:lt_att}
\begin{aligned}
    &\mathrm{\mathbf{Att}}^\mathrm{LT}_l(X_l^{t},X_l^{{[m]}}|E^{{[m]}}) = \\
    &\mathrm{\mathbf{Att}}_l(X_l^{t} W^{key}_l, X_l^{{[m]}} W^{key}_l, X_l^{{[m]}} W^{value}_l + E^{{[m]}}),
\end{aligned}
\end{equation}
where $X_l^{{[m]}}=\mathrm{\mathbf{Concat}}(X_l^{m_1},...,X_l^{m_{M}})$ and $E^{{[m]}}=\mathrm{\mathbf{Concat}}(E^{m_1},...,E^{m_M})$ are the input feature embeddings and ID embeddings of memory frames with indices ${[m]}=\{m_1,...,m_M\}$. Moreover, $W^{key}_l \in \mathbb{R}^{d_c \times d_{key}}$ and $W^{value}_l\in \mathbb{R}^{d_c \times d_{value}}$ are trainable parameters of the space projections for feature matching and value propagation, respectively. Instead of employing different projections for $X_l^{t}$ and $X_l^{{[m]}}$, the LSTT training shows more stability with a siamese-like matching strategy, which means matching the features within the same embedding space ($l$-th features with the same projection of $W^{key}_l$).

\noindent\textbf{Short-Term Attention} is further utilized to aggregate information in a spatial-temporal neighborhood for each location in the current frame. As frame motions or alterations across sequential video frames are generally smooth and uninterrupted, object matching and propagation between contiguous frames can be processed within a small spatial-temporal proximity, resulting in better efficacy and noise robustness than non-local processes. 

Assuming there are $n_\mathrm{ST}$ neighboring frames with indices $\mathbf{n_\mathrm{ST}}(t)=\{t-1,...,t-n_\mathrm{ST}\}$ in the spatial-temporal proximity, the features and ID embeddings of these frames are $X_l^{\mathbf{n_\mathrm{ST}}(t)}=\mathrm{\mathbf{Concat}}(X_l^{t-1},...,X_l^{t-n_\mathrm{ST}})$ and $E^{\mathbf{n_\mathrm{ST}}(t)}=\mathrm{\mathbf{Concat}}(E^{t-1},...,E^{t-n_\mathrm{ST}})$. Consequently, the formula for the short-term attention at each spatial location $p$ of the current frame is,
\begin{equation}\small\label{eq:st_att}
\begin{aligned}
    &\mathrm{\mathbf{Att}}^\mathrm{ST}_l(X_l^{t},X_l^{\mathbf{n_\mathrm{ST}}(t)}|E^{\mathbf{n_\mathrm{ST}}(t)}, p) = \\
    &\mathrm{\mathbf{Att}}^\mathrm{LT}_l(X_{l}^{t,p},X_{l}^{\mathbf{n_\mathrm{ST}}(t),\mathcal{N}(p)}|E_{l}^{\mathbf{n_\mathrm{ST}}(t),\mathcal{N}(p)}),
\end{aligned}
\end{equation}
where $X_{l}^{t,p} \in \mathbb{R}^{1 \times d_c} $ is the feature of $X_{l}^{t}$ at location $p$, $\mathcal{N}(p)$ is a $\lambda \times \lambda$ spatial neighborhood centered at location $p$, and thus $X_{l}^{\mathbf{n_\mathrm{ST}}(t),\mathcal{N}(p)}$ and $E_{l}^{\mathbf{n_\mathrm{ST}}(t),\mathcal{N}(p)}$ are the features and ID embeddings in the spatial-temporal proximity, respectively, with a shape of $n_\mathrm{ST}\lambda^2\times d_c$. 

When extracting features from the first frame, there are no memory frames or previous frames, we use $X_l^{1}$ to replace $X_l^{{[m]}}$ and $X_l^{\mathbf{n_\mathrm{ST}}(t)}$. Therefore, long-term attention and short-term attention are transformed into forms like self-attention without altering the network structures and parameters. 

\noindent\textbf{Offline Scalable AOT.} Based on LSTT, we can introduce the AOT framework with offline scalability. Unlike previous methods (\eg,~\cite{spacetime,cfbi,cheng2021stcn}), which have unscalable architecture, we make it easy to design AOT models with various levels of accuracy and speed by adjusting the depth of S-LSTT. An AOT model equipped with $n_l$ layers of S-LSTT is trained using the sequential training strategy~\cite{cfbi,cfbip}. For each training frame sample, we formulate the loss function of AOT as follows, where $Y'$ and $Y$ represent the prediction and ground-truth segmentation, respectively:
\begin{equation}\small
\label{eq:aot_loss}
  \mathcal{L}_\mathrm{AOT}(Y',Y)=\beta \mathcal{L}_\mathrm{BCE}(Y',Y) + (1 - \beta) \mathcal{L}_\mathrm{SJ}(Y',Y),
\end{equation}
where $\mathcal{L}_\mathrm{BCE}$ and $\mathcal{L}_\mathrm{SJ}$ denote bootstrapped cross-entropy loss and soft Jaccard loss~\cite{nowozin2014optimal}, respectively. The hyperparameter $\beta$ is used to balance these two losses.

\vspace{-3mm}
\subsection{Associating Objects with Scalable Transformers}\label{sec:aost}
\subsubsection{Scalable Long Short-Term Transformer (S-LSTT)}\label{sec:SLSTT}

\new{To further offer online scalability for VOS, we carefully craft a scalable version of LSTT, \ie, S-LSTT.}

\noindent\textbf{S-LSTT} stacks a series of S-LSTT blocks, but we can construct sub-transformers of S-LSTT with different depths. \new{In an S-LSTT with $n_l$ layers (\eg, 3 layers) at most, the sub-transformer of S-LSTT has a variable depth $n_l'$, which can be changed from $1$ to $n_l$, and a $n_l'$-layer (\eg, 2-layer) sub-transformer shares the parameters of S-LSTT's first $n_l'$ layers (\eg, first 2 in 3 layers)}, and AOST's encoder and decoder are shared for all sub-transformers. During inference, it is simple to alternate between different sub-transformers within the S-LSTT architecture. This allows us to select a deeper sub-transformer for superior accuracy or a shallower one for quicker processing speed. 

S-LSTT follows a similar architecture of our LSTT, but all the long-term and short-term attention layers are equipped with newly proposed layer-wise ID-based attention (Eq.~\ref{equ:id_v2}) for achieving better representative ability.


\begin{figure}[t!]
\begin{center}

\begin{subfigure}[b]{.43\columnwidth}
			\centering
			\includegraphics[height=4cm]{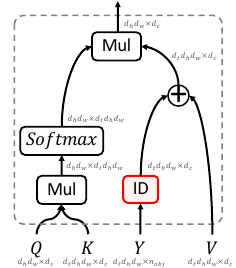}
   \vspace{-1mm}
			\caption{ID-based Att.}\label{fig:block}
\end{subfigure}
\begin{subfigure}[b]{.52\columnwidth}
			\centering
			\includegraphics[height=4cm]{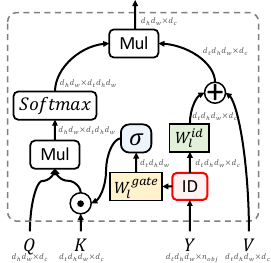}
   \vspace{-1mm}
			\caption{Layer-wise ID-based Att.}\label{fig:block_v2}
\end{subfigure}

\end{center}
\vspace{-4mm}
\caption{\textbf{Layer-wise ID-based Attention (\S\ref{sec:id_att})} is improved with trainable identification weight $W^{id}_{l}$ and gate weight $W^{gate}_{l}$, compared to the vanilla ID-based attention (Eq.\ref{eq: attid})}\label{fig:blocks}
\vspace{-3.0mm}
\end{figure}

\vspace{-2mm}
\subsubsection{Layer-wise ID-based Attention}\label{sec:id_att}
Although the ID-based attention, Eq.~\ref{eq: attid}, provides a straightforward method to match and propagate ID embeddings in a single attention layer, we discovered that the representation ability of value embeddings is limited when stacking multiple layers of Eq.~\ref{eq: attid}. Simply put, the uniformity of the ID embedding across all layers causes all value embeddings $V$ to align in a single embedding space. Consequently, the value embeddings lose their diversity and representation capabilities. In addition, since ID embeddings are merely appended to value embeddings $V$, key embeddings $K$ cannot leverage identification information - such as the location of background regions - to enhance matching processes.

Hence, we propose to construct and utilize identification in different embedding spaces for different layers. Let $W^{id}_{l}\in \mathbb{R}^{d_c \times d_c}$ and $W^{gate}_{l}\in \mathbb{R}^{d_c \times 1}$ denote trainable layer-wise \textbf{Identification Weight} and \textbf{Gate Weight} for $l$-th attention layer, we modify Eq.~\ref{eq: attid} to be:
\begin{equation}\small
\label{equ:id_v2}
\begin{aligned}
    V'&=\mathrm{\mathbf{AttID}}_l(Q,K,V|E,W^{id}_{l},W^{gate}_{l}) \\
    &=\mathrm{\mathbf{Att}}(Q,K\odot \sigma(EW^{gate}_{l}),V+EW^{id}_{l}),
\end{aligned}
\end{equation}
where $E=\mathrm{\mathbf{ID}}(Y,D)$ as Eq.~\ref{equ:id}, $\odot$ denotes Hadamard product, and $\sigma(\cdot)$ is a gate activation function~\cite{gct}. We set $\sigma(\cdot)=1+tanh(\cdot)$, which can learn a gate of identify mapping and improves training stability~\cite{gct}. Fig.~\ref{fig:block_v2} shows an illustration of layer-wise ID-based attention.


By introducing the identification weight $W^{id}{l}$, the identification embedding $E$ is projected into multiple spaces for varying attention layers before it is added to $V$, improving the representation ability of the feature space. Moreover, a gate function $\sigma(W^{gate}_{l}E)$, which is reliant on the identification information, further spatially adjusts the key embedding $K$ to focus on object regions. The gate weight $W^{gate}_{l}$ lowers the channel dimensions of $E$ to a single-channel attention map, and the gate function modifies $K$ in a lightweight position-wise manner.


\noindent\textbf{Layer-wise Long-Term Attention}. Based on the layer-wise ID-based attention (Eq.~\ref{eq: attid}), we can modify the long-term attention (Eq.~\ref{eq:lt_att}) into a layer-wise ID-based version:
\begin{equation}\small
\begin{aligned}
    &\mathrm{\mathbf{AttID}}^\mathrm{LT}_l(X_l^{t},X_l^{{[m]}}|E^{{[m]}}) = \\
    &\mathrm{\mathbf{AttID}}_l(X_l^{t} W^{key}_l, X_l^{{[m]}} W^{key}_l, X_l^{{[m]}} W^{value}_l|E^{{[m]}},W^{id}_{l},W^{gate}_{l}).
\end{aligned}
\end{equation}

\noindent\textbf{Layer-wise Short-Term Attention} can also be formulated based on the above modified long-term attention as:
\begin{equation}\small
\begin{aligned}
    &\mathrm{\mathbf{AttID}}^\mathrm{ST}_l(X_l^{t},X_l^{\mathbf{n_\mathrm{ST}}(t)}|E^{\mathbf{n_\mathrm{ST}}(t)}, p) = \\
    &\mathrm{\mathbf{AttID}}^\mathrm{LT}_l(X_{l}^{t,p},X_{l}^{\mathbf{n_\mathrm{ST}}(t),\mathcal{N}(p)}|E_{l}^{\mathbf{n_\mathrm{ST}}(t),\mathcal{N}(p)}),
\end{aligned}
\end{equation}
which calculating within a spatial-temporal proximity $\mathbf{n_\mathrm{ST}}(t)$ following the vanilla short-term attention (Eq.~\ref{eq:st_att}).

\vspace{-2mm}
\subsubsection{Scalable Supervision}\label{sec:scalable_supervision}
To achieve online scalability, we propose the Scalable Supervision strategy to train our AOST framework efficiently. Our AOST model, equipped with S-LSTT with $n_l$ layers, is switchable among $n_l$ sub-networks. Unlike AOT and previous VOS methods which train each model individually, the layer-wise designs (Eq.~\ref{equ:id_v2}) of ID-based attention allows us to train all sub-networks simultaneously and effectively. We denote the prediction of the sub-network with $n_l'$ S-LSTT layers as $Y'_{n_l'}$. The loss formula under scalable supervision for AOST is formulated as follows:

\begin{equation}\small
\label{eq:scalable_loss}
  \mathcal{L}_\mathrm{AOST}=\frac{\sum_{n_l'=1}^{n_l}\alpha^{n_l'}\mathcal{L}_\mathrm{AOT}(Y'_{n_l'},Y)}{\sum_{n_l'=1}^{n_l}\alpha^{n_l'}},
\end{equation}
Here, it is important to note that $n_l'$ does not serve as an index, but rather as the exponent used in conjunction with $\alpha$, which represents a scaling weight that re-weights the loss ratio of various sub-networks. Typically, shallow sub-networks exhibit lower accuracy, higher losses, and larger training gradients. Therefore, an $\alpha$ value greater than $1$ will emphasize the deeper sub-networks' losses, helping to balance the gradients of sub-networks. By selecting a suitable $\alpha$ value, we can maintain the performance of deeper sub-networks. In our default setting, $\alpha$ is set to 2. Additionally, scalable supervision can optimize parameter utilization and decrease computational costs by allowing various sub-networks to share the same encoder feature and reuse the S-LSTT feature of the deepest sub-network with $n_l$ layers.

\vspace{-1.5mm}

\subsection{Implementation Details}\label{sec:implementation}




\noindent\textbf{AOT Variants:} To validate the effectiveness of our identification mechanism and hierarchical LSTT, we build AOT~\cite{aot} variants with different LSTT layer number $n_l$ or long-term memory size ${[m]}$ for comparisons: 
\begin{itemize}[leftmargin=*, noitemsep, topsep=0pt]
\item  \textbf{AOT-L}arge:\hspace{0.05em} $n_l=3$, ${[m]}=\{1,1+\delta,1+2\delta,...\}$; 
\item  \textbf{AOT-B}ase:\hspace{0.5em} $n_l=3$, ${[m]}=\{1\}$; 
\item  \textbf{AOT-S}mall: $n_l=2$, ${[m]}=\{1\}$; 
\end{itemize}

\new{In AOT-S/B/L, only the first frame is considered in long-term memory, similar to \cite{feelvos,cfbi}, leading to stable run-time speeds.} In AOT-L, the predicted frames are stored in long-term memory per $\delta$ frames, following the commonly-used memory reading strategy~\cite{spacetime} with better performance. 

\noindent\textbf{AOST Variants:} To evaluate the layer-wise ID-based attention and developing online scalability, we further build AOST variants whose hyper-parameters are: 
\begin{itemize}[leftmargin=*, noitemsep, topsep=0pt]
\item  \textbf{AOST}$^{n_l'=3}$: $n_l=3$, $n_l'=3$, ${[m]}=\{1,1+\delta,1+2\delta,...\}$; 
\item  \textbf{AOST}$^{n_l'=2}$: $n_l=3$, $n_l'=2$, ${[m]}=\{1,1+\delta,1+2\delta,...\}$; 
\item  \textbf{AOST}$^{n_l'=1}$: $n_l=3$, $n_l'=1$, ${[m]}=\{1,1+\delta,1+2\delta,...\}$.
\end{itemize}

Following AOT-L, all the AOST variants with different depths ($n_l=1/2/3$) of sub-networks share a base AOST model with 3 layers of S-LSTT and employ the memory reading strategy. We set $\delta$ to 2/5 for training/testing of AOST variants and AOT-L.

\noindent\textbf{Network Details:} To sufficiently validate the effectiveness, we use light-weight backbone encoder, MobileNet-V2~\cite{sandler2018mobilenetv2}, and decoder, FPN~\cite{fpn} with GroupNorm~\cite{gn} in default. To verify robustness, we also use stronger ResNet-50 (R50)~\cite{resnet} and Swin-B~\cite{swin} as the encoder. The spatial neighborhood size $\lambda$ is set to 15, and the size of the identity bank, $M$, is set to 10, which is consistent with the maximum object number in the benchmarks~\cite{youtubevos,davis2017}. \new{For background regions, we assign a unique learnable identity}.

For MobileNet-V2 encoder, we increase the final resolution of the encoder to $1/16$ by adding a dilation to the last stage and removing a stride from the first convolution of this stage. For ResNet-50 and SwinB encoders, we remove the last stage directly as~\cite{spacetime}. The encoder features are flattened into sequences before S-LSTT. In S-LSTT, the feature dimension is 256, and the head number is 8 for all the attention modules. To increase the receptive field of S-LSTT, we insert a depth-wise convolution layer with a kernel size of 5 in the middle of each feed-forward module. The short-term memory only considers the previous ($t-1$) frame, similar to the local matching strategy~\cite{feelvos,cfbi}. After S-LSTT, all the output features of S-LSTT blocks are reshaped into 2D shapes and will serve as the decoder input. Then, the FPN decoder progressively increases the feature resolution from $1/16$ to $1/4$ and decreases the channel dimension from 256 to 128 before the final output layer.

\noindent\textbf{Training Details:} Following~\cite{rgmp,spacetime,EGMN,KMN}, the training stage is divided into two phases: (i) pre-training on synthetic video sequence generated from static image datasets~\cite{voc,coco,cheng2014global,shi2015hierarchical,semantic} by randomly applying multiple image augmentations~\cite{rgmp}. (ii) main training on the VOS benchmarks~\cite{youtubevos,davis2017} by randomly applying video augmentations~\cite{cfbi}. 

\begin{figure}[t!]
\centering
\includegraphics[width=0.98\columnwidth]{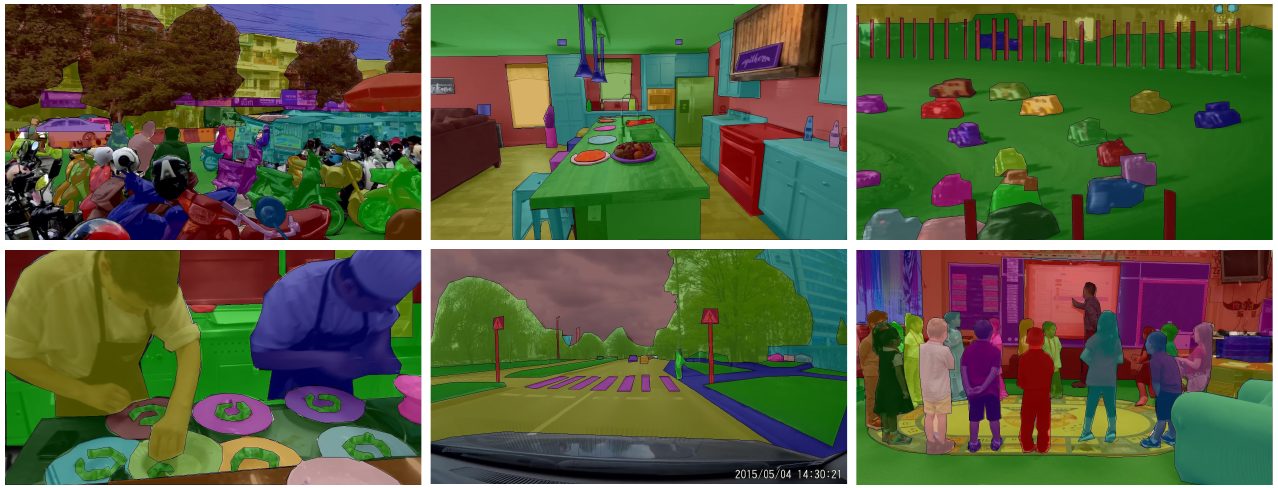}
\caption{\textbf{Examples of VOSW (\S\ref{sec:VOSW})}. VOSW covers a wide range of real-world categories and scenarios with challenges of dense similar objects (\eg, \textit{humans, vehicles, chairs, tablewares}), diverse object categories, complicated occlusions (\eg, occlusion between moving objects), etc. }\label{fig:vosw_example}\vspace{-4mm}
\end{figure}

All the videos are firstly down-sampled to 480p resolution, and the cropped window size is 465$\times$ 465. For optimization, we adopt the AdamW~\cite{adamw} optimizer and the sequential training strategy~\cite{cfbi}, whose sequence length is set to 5. The loss function is a 0.5:0.5 combination of bootstrapped cross-entropy loss and soft Jaccard loss~\cite{nowozin2014optimal}. For stabilizing the training, the statistics of BN~\cite{bn} modules and the first two stages in the encoder are frozen, and Exponential Moving Average (EMA)~\cite{polyak1992acceleration} is used. Besides, we apply stochastic depth~\cite{huang2016deep} to the self-attention and the feed-forward modules in S-LSTT.

The batch size is 16 and distributed to 4 Tesla V100 GPUs. For pre-training, we use an initial learning rate of $4\times10^{-4}$ and a weight decay of $0.03$ for 100,000 steps. For main training, the initial learning rate is $2\times10^{-4}$, the weight decay is $0.07$, and the training steps are 100,000/50,000 for YouTube-VOS/DAVIS. To relieve over-fitting, the initial learning rate of encoders is reduced to a 0.1 scale of other network parts. All the learning rates gradually decay to $2\times10^{-5}$ in a polynomial manner~\cite{cfbi}. 
1 Tesla V100 GPU is used for evaluation, and the scales used in test-time multi-scale augmentation are $\{1.2, 1.3, 1.4\}\times 480p$.


\vspace{-2.5mm}

\section{VOS in the Wild (VOSW) Dataset}\label{sec:VOSW}

\begin{figure}[t!]
\begin{center}
\includegraphics[width=0.99\columnwidth]{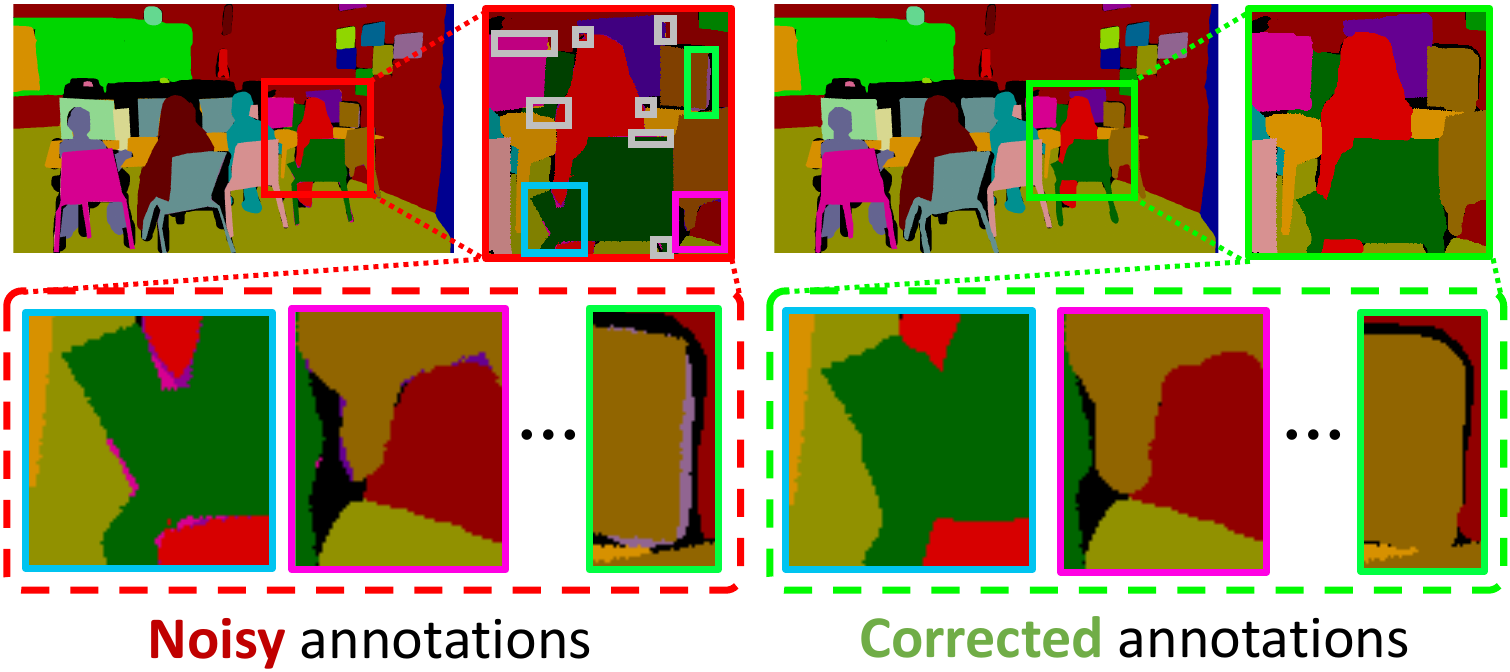}
\end{center}
\vspace{-4mm}
\caption{\textbf{Data clean in building VOSW (\S\ref{sec:VOSW})}. VIPSeg~\cite{miao2022large} has considerable noise at objects' edges. We carefully corrected all the noise to improve data quality and ensure VOSW's evaluation reliability.}\label{fig:vosw_noise}

\end{figure}
\begin{table}[t!]
\begin{center}
\caption{
\new{\textbf{Dataset Comparison (\S\ref{sec:VOSW})}. Compared to commonly used benchmarks~\cite{youtubevos,davis2017,davis2016}, VOSW yields richer semantic diversity and includes more objects per video. }
Obj/V: objects per video. T/V: time duration per video.}\label{tab:dataset_comparison}
\vspace{-0mm}
\setlength{\tabcolsep}{3.5pt}
\renewcommand\arraystretch{1.4}
\begin{tabular}{|l|c|c|c|c|}
\hline
\rowcolor[HTML]{FAFAFA} 
Dataset                             & Video                        & Class                       & Obj/V                        & T/V                                                                             \\ \hline
DAVIS 2016~\cite{davis2016} Train/Val                        & 30/20                            & -                             & 1.0                            & 2.9s                            \\ \hline
DAVIS 2017~\cite{davis2017} Train/Val/Test                        & 60/30/30                            & -                             & \textless3.0                            & \textless2.9s                           \\ \hline
YTB 2018~\cite{youtubevos} Train/Val                  & 3471/474                           & 94                            & 1.9                            & 4.4s                           \\ \hline
YTB 2019~\cite{youtubevos} Train/Val                  & 3471/{507} & 94                            & 2.1                            & 4.5s                           \\ \hline
\new{VOST}~\cite{tokmakov2023breaking}  & \new{713} & \new{\textbf{155}} & \new{2.3} & \new{21.2s} \\ \hline
\new{UVO}~\cite{wang2021unidentified}  & \new{1200} & \new{open} & \new{\underline{12.3}} & \new{3.0s} \\ \hline
\textbf{VOSW (Ours)} & 348                           & \underline{125} & \textbf{15.0} & {4.6s}              \\ \hline
\end{tabular}

\end{center}
\vspace{-3.5mm}
\end{table}

In numerous pivotal video application scenarios (\eg, sporting events, self-driving cars, and augmented reality), there exist numerous similar objects such as humans, vehicles, and other common objects. Therefore, precise multi-object VOS holds significant value and has attracted the attention of researchers. Although there are several popular multi-object benchmarks in the VOS domain (\eg, YouTube-VOS~\cite{youtubevos} and DAVIS~\cite{davis2017}), these datasets contain a limited average number of annotated objects per video ($<$ 3 object/video), failing to provide researchers with evaluation scenarios that include dense objects. To better address the challenges of VOS in real-world applications, we create a comprehensive evaluation benchmark for multi-object Video Object Segmentation in the Wild (VOSW) encompassing highly dense object annotations, diverse categories, and challenging scenarios.


To construct VOSW, we utilized the recently released VIPSeg dataset~\cite{miao2022large} as our primary data source. Designed for video panoptic segmentation (VPS), VIPSeg has released 3149 videos with corresponding segmentation annotations for 58 thing categories and 66 stuff categories. However, despite the density of the annotations, VIPSeg is not suitable for direct use as a VOS dataset due to task dissimilarities (VOS \vs VPS), noisy annotations, data distributions, and other related issues. For this reason, we carefully make several indispensable improvements to create VOSW:


\begin{figure}[t!]
    \centering
    \includegraphics[width=1\linewidth]{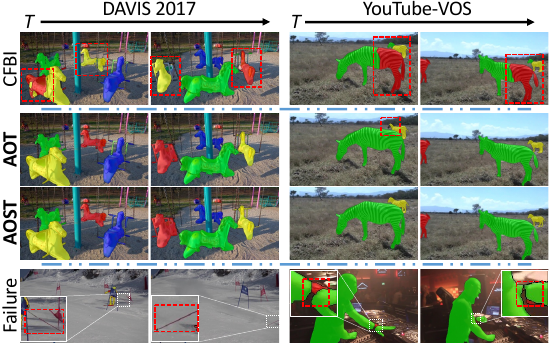}
    \caption{\textbf{Qualitative results (\S\ref{sec:compare}) on YouTube-VOS~\cite{youtubevos} and DAVIS~\cite{davis2017}}. (top) CFBI~\cite{cfbip,cfbi} performs worse AOST \& AOT when segmenting multiple highly similar objects (\textit{carousels} and \textit{zebras}). (middle) With S-LSTT, AOST$^{n_l'=3}$ produces finer results. (bottom) AOST/AOT fails to segment some tiny objects (\textit{ski poles} and \textit{watch}) since we have no specific design for processing rare tiny objects.}\vspace{-2.0mm}
    \label{fig:comparisons}
    
\end{figure}

\noindent\textbf{Data Collection}. The collection of videos from VIPSeg for building VOSW has been a meticulous process with the following objectives: \textit{(i) Diversity of Categories}: The VOSW dataset is constructed to cover all the thing and staff categories from VIPSeg, to ensure that there is no loss of category diversity. \textit{(ii) Dense Objects}: During video collection, we prioritize videos with a higher number of objects and complex camera motion to guarantee the challenge of VOSW. \textit{(iii) High Quality}: Unlike video panoptic segmentation, which focuses more on recognizing things and stuff, VOS places more emphasis on the robustness of tracking and segmentation quality. Thus, we manually filter videos with obvious tracking trajectory errors and imprecise mask annotations. After collecting videos based on these objectives, the VOSW benchmark now contains 348 videos.


\noindent\textbf{Data Clean}. Although the data collection strategy filters out some noise data, the selected videos may still have significant noise errors. Fig.~\ref{fig:vosw_noise} shows that VIPSeg has a considerable amount of noise present at the edges of objects in the annotations. This leads to two significant issues in VOS: \textbf{First}, it generates many noisy objects that can mislead algorithms' tracking process. \textbf{Second}, since VOS computes metrics individually per object and then takes an average, the presence of noisy objects will significantly lower the evaluation score and reliability. Hence, we manually corrected the noise errors affecting VOSW to improve the data quality. This correction ensured greater reliability of the evaluation benchmark as well. Post data collection and noise correction, the average number of objects in VOSW rose to 15.0 object/video, which is higher than VIPSeg's dataset containing noisy objects (13.3 object/video).


\noindent\textbf{Re-annotation}. To comply with the task setting of VOS, we made necessary adjustments to the format of all annotations. \textbf{First}, thing and staff annotations are converted into object annotations. In this process, we assigned a different index to each instance in thing classes or semantic regions in stuff classes, after which we shuffled their orders. \textbf{Second}, we annotated the first appearance frame of each object in the video. During the evaluation, VOSW provides masks in the first appearance frame of every object. Objects may appear in different frames instead of only the $t$=1 frame.


\begin{figure}[t!]
    \centering
    \vspace{-0.0mm}
    \includegraphics[width=1\linewidth]{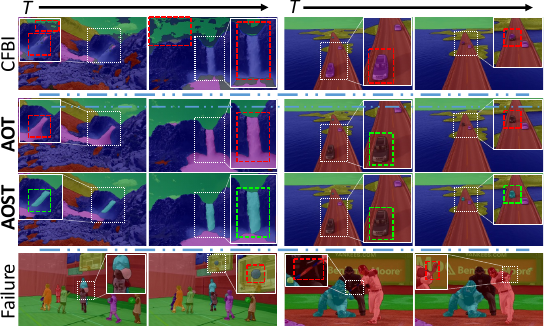}
    \caption{\textbf{Qualitative results (\S\ref{sec:compare}) on our VOSW}. (top) STCN~\cite{cheng2021stcn} fails to track and segment moving objects (\textit{waterfall}, \textit{sky}, \textit{vehicle}) when tracking numerous targets. (middle) With the identification mechanism, AOST/AOT performs better than STCN in scenarios with multiple objects. Besides, AOST outperforms AOT when tracking highly similar objects. (bottom) AOST/AOT sometimes fails to segment tiny objects (\textit{basketball}) or highly deformed objects (\textit{baseball bat}).}
    \label{fig:vosw_comparisons}
    \vspace{-3.5mm}
\end{figure}
Fig.~\ref{fig:vosw_example} shows that VOSW contains challenging scenes with numerous similar targets, intricate dense annotations, and complicated occlusions, etc. Based on Table~\ref{tab:dataset_comparison}, the average number of objects in VOSW is considerably higher ($>$ 5$\times$) than YouTube-VOS and DAVIS, rendering it more fitting for evaluating the performance of VOS algorithms in scenarios with dense targets. Additionally, VOSW presents greater category diversity, which can better reflect the robustness of VOS algorithms during cross-dataset evaluation.

\vspace{-2.5mm}


\section{Experimental Results}\label{sec:experiments}
\noindent\textbf{Datasets.} We evaluate AOT/AOST variants on our \textbf{VOSW} (\S\ref{sec:VOSW}) and commonly-used benchmarks:
\begin{itemize}[leftmargin=*, noitemsep, topsep=0pt]
\item \textbf{YouTube-VOS}~\cite{youtubevos} are large-scale benchmarks for multi-object VOS, including 3,471 videos for training and 474/507 videos in the 2018/2019 validation (Val) split. The frames and annotations are provided at 6FPS, and all-frame (30FPS) video frames are also supplied.

\item \textbf{DAVIS 2017}~\cite{davis2017} contains 60/30 multi-object videos in the Train/Val split. Also, It provides a Test set that comprises 30 videos with more challenging scenarios.

\item \textbf{DAVIS 2016}~\cite{davis2016} has 50 single-object videos with finely detailed labeling at 24FPS and is partitioned into the Train/Val split with 30/20 videos.
\end{itemize}

\begin{table}[t!]
	\centering
	\caption{\textbf{Quantitative results (\S\ref{sec:compare}) on YouTube-VOS 2018}~\cite{youtubevos}. 
	AF: using \textit{All-Frame} (30FPS) videos instead of the default (6FPS) videos.
	  $^\dagger$: setting $\alpha=\infty$ during training. The 1st and 2nd results are marked with \textbf{bold} and \underline{underline}.
	}\label{tab:youtubevos}
	\vspace{-3.5mm}
\centering
\setlength{\tabcolsep}{4.5pt}
\renewcommand\arraystretch{1.1}
\begin{tabular}{rccccccc}
\bottomrule[1pt]\rowcolor[HTML]{FAFAFA}
                                         &            &                            & \multicolumn{2}{c}{Seen}      & \multicolumn{2}{c}{Unseen}    &               \\ \hline\rowcolor[HTML]{FAFAFA}
Method                                  & AF         & $\mathcal{J}\&\mathcal{F}$ & $\mathcal{J}$ & $\mathcal{F}$ & $\mathcal{J}$ & $\mathcal{F}$ & FPS           \\ \toprule[0.8pt]
STM\hspace{0.68em}\hspace{0.45em}\pub{ICCV19}\hspace{0.48em}\hspace{0.77em}\cite{spacetime}         & \checkmark & 79.4                       & 79.7          & 84.2          & 72.8          & 80.9          & -             \\ 
KMN\hspace{0.68em}~\pub{ECCV20}\hspace{0.77em}\hspace{0.48em}\cite{KMN}               & \checkmark & 81.4                       & 81.4          & 85.6          & 75.3          & 83.3          & -             \\
EGMN\hspace{0.68em}~\pub{ECCV20}\hspace{0.77em}\hspace{0.48em}\cite{EGMN}                                      & \checkmark & 80.2                       & 80.7          & 85.1          & 74.0          & 80.9          & 7.7           \\
CFBI\hspace{0.68em}~\pub{ECCV20}\hspace{0.77em}\hspace{0.48em}\cite{cfbi}             &            & 81.4                       & 81.1          & 85.8          & 75.3          & 83.4          & 3.4           \\
LWL\hspace{0.68em}~\pub{ECCV20}\hspace{0.77em}\cite{LWLVOS}            & \checkmark & 81.5                       & 80.4          & 84.9          & 76.4          & 84.4          & -             \\

STCN~\pub{NeurIPS21}\hspace{0.77em}\hspace{0.48em}\cite{cheng2021stcn} & \checkmark & 83.0                       & 81.9          & 86.5          & 77.9          & 85.7          & 8.4  \\
Joint\hspace{0.68em}~\pub{CVPR21}\hspace{0.77em}\cite{mao2021joint}                                  & \checkmark & 83.1                       & 81.5          & 85.9          & 78.7          & 86.5          & 10.0          \\
RPCM\hspace{0.68em}~\pub{AAAI22}\hspace{0.77em}\cite{rpcm}             &            & 84.0                       & 83.1          & 87.7          & 78.5          & 86.7          & -             \\
PCVOS\hspace{0.68em}~\pub{CVPR22}~\cite{pcvos_cvpr22}                                     & \checkmark & 84.6                       & 83.0          & 88.0          & 79.6          & 87.9          & 10.5          \\
\new{XMem}\hspace{0.68em}~\pub{ECCV22}\hspace{0.77em}\cite{cheng2022xmem}                                       & \checkmark & \new{85.7}                       & \new{84.6}          & \new{89.3}          & \new{80.2}          & \new{88.7}          & \new{22.6} \\

\hline
\multicolumn{8}{c}{\textit{\new{w/ Hierarchical Structure}}} \\
\hline
SST\hspace{0.68em}~\pub{CVPR21}~\cite{sstvos}            &            & 81.7                       & 81.2          & -             & 76.0          & -             & -             \\
HMMN\hspace{0.68em}\hspace{0.45em}\pub{ICCV21}~\cite{hmmn}             & \checkmark & 82.6                       & 82.1          & 87.0          & 76.8          & 84.6          & -             \\
CFBI+\hspace{0.68em}\hspace{0.35em}\pub{PAMI21}\hspace{0.48em}\hspace{0.77em}\cite{cfbip}          &            & 82.8                       & 81.8          & 86.6          & 77.1          & 85.6          & 4.0           \\
DeAOT\hspace{0.1em}~\pub{NeurIPS22}\hspace{0.77em}\cite{yang2022decoupling} &  & \textbf{86.0}                       & \underline{84.9}          & 89.9          & \textbf{80.4}          & \textbf{88.7}          & \underline{22.4}  \\
\hline\hline
\rowcolor[HTML]{F9FFF9}
\rowcolor[HTML]{F9FFF9}
AOT-S~\pub{NeurIPS21}\hspace{0.77em}\cite{aot}                                    &            & 82.6                       & 82.0          & 86.7          & 76.6          & 85.0          & \textbf{27.1}        \\
\rowcolor[HTML]{F9FFF9}
AOT-B~\pub{NeurIPS21}\hspace{0.77em}\cite{aot}                                    &            & 83.5                       & {82.6}        & 87.5          & {77.7}        & 86.0          & {20.5}          \\ 
\rowcolor[HTML]{F9FFF9}
AOT-L~\pub{NeurIPS21}\hspace{0.77em}\cite{aot}                                    &            & {83.8}                     & {82.9}        & {87.9}        & {77.7}        & {86.5}        & 16.0          \\ \hdashline
\rowcolor[HTML]{F9FBFF}
AOST$^{n_l'=3}$\hspace{0.5em}                            &            & 84.4                       & 83.5          & 88.5          & 78.4          & 87.0          & 15.9          \\
\rowcolor[HTML]{F9FBFF}
R50-AOST$^{n_l'=3}$\hspace{0.5em}                        &            & 85.0                       & 83.8          & 88.8          & {79.3}          & {87.9}          & 14.9          \\
\rowcolor[HTML]{F9FBFF}
R50-AOST$^{n_l'=3}$$^\dagger$                               &            & {84.8}                     & {84.1}        & {88.9}        & {78.8}        & {87.3}        & 14.9          \\
\rowcolor[HTML]{F9FBFF}
R50-AOST$^{n_l'=3}$$^\dagger$                               & \checkmark & {85.4}                     & \textbf{85.1} & \textbf{90.2} & {78.9}        & {87.3}        & 6.3           \\ 
\rowcolor[HTML]{F9FBFF}
SwinB-AOST$^{n_l'=3}$\hspace{0.5em}                     &            & 84.9                       & {84.5}          & 89.6          & 78.6          & 87.0          & 9.3           \\ 
\rowcolor[HTML]{F9FBFF}
SwinB-AOST$^{n_l'=3}$$^\dagger$                             &            & {85.1}              & {84.5} & {89.4} & {79.0} & {87.5} & 9.3           \\
\rowcolor[HTML]{F9FBFF}
SwinB-AOST$^{n_l'=3}$$^\dagger$                             & \checkmark & \underline{85.8}              & \textbf{85.1} & \underline{90.1}        & \underline{79.6} & \underline{88.2} & 5.1           \\ 
\toprule[1pt]
\end{tabular}
\vspace{-4.0mm}
\end{table}

All the variants are trained on the training splits of YouTube-VOS 2019 and DAVIS-2017 in the default setting. To validate our generalization ability, we choose only the last checkpoint of each variant, and all the benchmarks share the same model parameters. When evaluating VOSW/YouTube-VOS, we use the default 5FPS/6FPS videos, with a resolution constraint of less than $1.3 \times480p$. As to DAVIS, standard 480p 24FPS videos are used.

\noindent\textbf{Metrics.} The evaluation metrics encompass:
\begin{itemize}[leftmargin=*, noitemsep, topsep=0pt]
\item $\mathcal{J}$ \textbf{Score}: as the Intersect over Union (IoU) score between the prediction and the ground truth mask.
\item $\mathcal{F}$ \textbf{Score}: as the boundary similarity measure between the boundary of the prediction and the ground truth.
\item $\mathcal{J}$\&$\mathcal{F}$ \textbf{Score}: as the average score of $\mathcal{J}$ and $\mathcal{F}$ scores.
\end{itemize}

\begin{table}[t!]
	\centering
	\caption{\textbf{Quantitative results (\S\ref{sec:compare}) on YouTube-VOS 2019}~\cite{youtubevos}. See Table~\ref{tab:youtubevos} for the definition of abbreviations.
	}\label{tab:youtubevos19}
	\vspace{-3.5mm}
\centering
\setlength{\tabcolsep}{4.5pt}
\renewcommand\arraystretch{1.1}
\begin{tabular}{rccccccc}
\bottomrule[1pt]\rowcolor[HTML]{FAFAFA}
                                         &            &                            & \multicolumn{2}{c}{Seen}      & \multicolumn{2}{c}{Unseen}    &               \\ \hline\rowcolor[HTML]{FAFAFA}
Method                                  & AF         & $\mathcal{J}\&\mathcal{F}$ & $\mathcal{J}$ & $\mathcal{F}$ & $\mathcal{J}$ & $\mathcal{F}$ & FPS                                                                                                                                                                \\ \toprule[0.8pt]
LWL\hspace{0.68em}~\pub{ECCV20}\hspace{0.77em}\cite{LWLVOS}            & \checkmark & 81.0                       & 79.6          & 83.8          & 76.4          & 84.2          & -             \\
CFBI\hspace{0.68em}~\pub{ECCV20}\hspace{0.48em}\hspace{0.77em}\cite{cfbi}             &            & 81.0                       & {80.6}        & {85.1}        & {75.2}        & {83.0}        & 3.4           \\

STCN~\pub{NeurIPS21}\hspace{0.77em}\hspace{0.48em}\cite{cheng2021stcn} & \checkmark & 82.7                       & 81.1          & 85.4          & 78.2          & 85.9          & 8.4   \\
RPCM\hspace{0.68em}~\pub{AAAI22}\hspace{0.77em}\cite{rpcm}             &            & 83.9                       & 82.6          & 86.9          & 79.1          & 87.1          & -             \\
PCVOS\hspace{0.68em}~\pub{CVPR22}~\cite{pcvos_cvpr22}                                     & \checkmark & 84.6                       & 83.0          & 88.0          & 79.6          & 87.9          & 10.5          \\
\new{XMem}\hspace{0.68em}~\pub{ECCV22}\hspace{0.77em}\cite{cheng2022xmem}                                       & \checkmark                        & \new{85.5}          & \new{84.8}    & \new{89.2}      & \new{80.3}          & \new{88.8}          & \new{22.6} \\
\hline
\multicolumn{8}{c}{\textit{\new{w/ Hierarchical Structure}}} \\
\hline
SST\hspace{0.68em}~\pub{CVPR21}~\cite{sstvos}            &           & 81.8                       & 80.9          & -             & 76.6          & -             & -             \\
HMMN\hspace{0.68em}\hspace{0.45em}\pub{ICCV21}~\cite{hmmn}             & \checkmark & 82.5                       & 81.7          & 86.1          & 77.3          & 85.0          & -             \\
CFBI+\hspace{0.68em}\hspace{0.35em}\pub{PAMI21}\hspace{0.77em}\hspace{0.48em}\cite{cfbip}          &            & 82.6                       & 81.7          & 86.2          & 77.1          & 85.2          & 4.0           \\
DeAOT\hspace{0.1em}~\pub{NeurIPS22}\hspace{0.77em}\cite{yang2022decoupling} &  & \underline{85.9}                       & 84.6          & 89.4          & \textbf{80.8}          & \textbf{88.9}          & \underline{22.4}  \\
\hline\hline
\rowcolor[HTML]{F9FFF9}
\rowcolor[HTML]{F9FFF9}
AOT-S~\pub{NeurIPS21}\hspace{0.77em}\cite{aot}                                    &            & 82.2                       & 81.3          & 85.9          & 76.6          & 84.9          & \textbf{27.1}        \\
\rowcolor[HTML]{F9FFF9}
AOT-B~\pub{NeurIPS21}\hspace{0.77em}\cite{aot}                                    &            & 83.3                       & {82.4}        & {87.1}        & 77.8          & 86.0          & {20.5}          \\
\rowcolor[HTML]{F9FFF9}
AOT-L~\pub{NeurIPS21}\hspace{0.77em}\cite{aot}                                    &            & {83.7}                     & {82.8}        & {87.5}        & {78.0}        & {86.7}        & 16.0          \\ \hdashline
\rowcolor[HTML]{F9FBFF}
AOST$^{n_l'=3}$\hspace{0.5em}                           &            & 84.4                       & 83.3          & 88.0          & 79.0          & 87.4          & {15.9}        \\
\rowcolor[HTML]{F9FBFF}
R50-AOST$^{n_l'=3}$\hspace{0.5em}                       &            & 84.9                       & 83.8          & 88.7          & 79.3          & 87.7          & {14.9}        \\
\rowcolor[HTML]{F9FBFF}
R50-AOST$^{n_l'=3}$$^\dagger$                               &            & {85.0}                     & {84.0}        & {88.7}        & {79.5}        & {87.9}        & 14.9          \\ 
\rowcolor[HTML]{F9FBFF}
R50-AOST$^{n_l'=3}$$^\dagger$                               & \checkmark & {85.3}                     & \underline{84.7}        & \underline{89.6}        & {79.2}        & {87.6}        & 6.3           \\
\rowcolor[HTML]{F9FBFF}
SwinB-AOST$^{n_l'=3}$\hspace{0.5em}                     &            & 85.0                       & 84.5          & 89.3          & 79.0          & 87.3          & {9.3}         \\
\rowcolor[HTML]{F9FBFF}
SwinB-AOST$^{n_l'=3}$$^\dagger$                             &            & {85.2}              & {84.2} & {88.9} & {79.8} & {88.0} & 9.3           \\
\rowcolor[HTML]{F9FBFF}
SwinB-AOST$^{n_l'=3}$$^\dagger$                             & \checkmark & \textbf{86.0}              & \textbf{85.0} & \textbf{89.8} & \underline{80.3} & \underline{88.8} & 5.1  \\ 
\toprule[1pt]
\end{tabular}
\vspace{-4.0mm}

\end{table}

\subsection{Compare with State-of-the-art Competitors}\label{sec:compare}

\noindent \textbf{YouTube-VOS 2018 \& 2019 Val}~\cite{youtubevos}. \new{As shown in Table~\ref{tab:youtubevos}, AOST/AOT variants achieve superior performance on YouTube-VOS 2018 Val compared to state-of-the-art competitors. With our identification mechanism, AOT-S (\textbf{82.6\%} $\mathcal{J}\&\mathcal{F}$) is comparable with STCN~\cite{cheng2021stcn} (83.0\%) while running about $3.2\times$ faster (27.1 \vs 8.4FPS). By further employing S-LSTT, our R50-AOST (\textbf{85.0\%}, \textbf{14.9FPS}) is more efficient and effective than recent competitors, \eg, PCVOS~\cite{pcvos_cvpr22} ({84.6\%}, {10.5FPS}), without using videos with higher frame rate. Setting $\alpha=\infty$ will sacrifice the scalability of AOST but slightly improves the performance of AOST$^{n_l'=3}$. By utilizing all-frame (30FPS) videos following~\cite{spacetime,cheng2021stcn,pcvos_cvpr22,llb_aaai23}, we can further boost the performance to \textbf{85.4\%}. Furthermore, we can improve the performance by using a stronger Swin-Base~\cite{swin} backbone (SwinB-AOST \textbf{85.8\%} \vs  R50-AOST 85.4\%). The significance of AOST/AOT variants is consistent on YouTube-VOS 2019 Val (Table~\ref{tab:youtubevos19}).}




\noindent \textbf{DAVIS 2017 Val \& Test}~\cite{davis2017}. Table~\ref{tab:davis} shows that our R50-AOST surpasses all the competitors on both the Val (\textbf{85.6\%}) and Test (\textbf{79.9\%}) splits and maintains an efficient speed (\textbf{17.5FPS}). Notably, such a multi-object speed is the same as our single-object speed on DAVIS 2016 (Table~\ref{tab:davis2016}). For the first time, processing multiple objects can be as efficient as processing a single one with our identification mechanism. Apart from this, Table~\ref{tab:aot_aost_compare} shows that R50-AOST can also be adapted to real-time by reducing S-LSTT's layer number $n_l'$. On the challenging Test split, R50-AOST$^{n_l'=2}$ achieves \textbf{78.1\%} at \textbf{24.3FPS}, outperforming STCN~\cite{cheng2021stcn} (76.1\%, 19.5FPS).



\begin{table*}
\begin{center}
\caption{\textbf{Comparisons (\S\ref{sec:compare}) between our \colorbox[HTML]{F8FAFE}{AOST} and \colorbox[HTML]{F8FEF8}{AOT}~\cite{aot}} on YouTube-VOS 2018 \& 2019~\cite{youtubevos} and DAVIS 2016~\cite{davis2016} \& 2017~\cite{davis2017} using three different backbones. For each comparison (with the same backbone), AOST$^{n_l'=1/2/3}$ shares parameters from the same model.}\label{tab:aot_aost_compare}
\vspace{-3.0mm}
\renewcommand\arraystretch{1.4}
\setlength{\tabcolsep}{4pt}
\begin{tabular}{|
>{\columncolor[HTML]{FAFAFA}}c 
>{\columncolor[HTML]{FAFAFA}}c 
>{\columncolor[HTML]{FAFAFA}}l |
>{\columncolor[HTML]{F9FFF9}}c 
>{\columncolor[HTML]{F9FBFF}}c 
>{\columncolor[HTML]{F9FBFF}}c 
>{\columncolor[HTML]{F9FBFF}}c |
>{\columncolor[HTML]{F9FFF9}}c 
>{\columncolor[HTML]{F9FBFF}}c 
>{\columncolor[HTML]{F9FBFF}}c 
>{\columncolor[HTML]{F9FBFF}}c |
>{\columncolor[HTML]{F9FFF9}}c 
>{\columncolor[HTML]{F9FBFF}}c 
>{\columncolor[HTML]{F9FBFF}}c 
>{\columncolor[HTML]{F9FBFF}}c |}
\hline
\multicolumn{3}{|c|}{\cellcolor[HTML]{FAFAFA}}                                                                                                                                                                     & \multicolumn{1}{c|}{\cellcolor[HTML]{FAFAFA}}                                               & \multicolumn{3}{c|}{\cellcolor[HTML]{FAFAFA}AOST}                                                                                                         & \multicolumn{1}{c|}{\cellcolor[HTML]{FAFAFA}}                                               & \multicolumn{3}{c|}{\cellcolor[HTML]{FAFAFA}AOST}                                                                                                         & \multicolumn{1}{c|}{\cellcolor[HTML]{FAFAFA}}                                               & \multicolumn{3}{c|}{\cellcolor[HTML]{FAFAFA}AOST}                                                                                                         \\ \cline{5-7} \cline{9-11} \cline{13-15} 
\multicolumn{3}{|c|}{\multirow{-2}{*}{\cellcolor[HTML]{FAFAFA}Method}}                                                                                                                                             & \multicolumn{1}{c|}{\multirow{-2}{*}{\cellcolor[HTML]{FAFAFA}\makecell{AOT-L\\\cite{aot}}}} & \multicolumn{1}{c|}{\cellcolor[HTML]{FAFAFA}${n_l'=3}$}    & \multicolumn{1}{c|}{\cellcolor[HTML]{FAFAFA}${n_l'=2}$} & \cellcolor[HTML]{FAFAFA}${n_l'=1}$ & \multicolumn{1}{c|}{\multirow{-2}{*}{\cellcolor[HTML]{FAFAFA}\makecell{AOT-L\\\cite{aot}}}} & \multicolumn{1}{c|}{\cellcolor[HTML]{FAFAFA}${n_l'=3}$}    & \multicolumn{1}{c|}{\cellcolor[HTML]{FAFAFA}${n_l'=2}$} & \cellcolor[HTML]{FAFAFA}${n_l'=1}$ & \multicolumn{1}{c|}{\multirow{-2}{*}{\cellcolor[HTML]{FAFAFA}\makecell{AOT-L\\\cite{aot}}}} & \multicolumn{1}{c|}{\cellcolor[HTML]{FAFAFA}${n_l'=3}$}    & \multicolumn{1}{c|}{\cellcolor[HTML]{FAFAFA}${n_l'=2}$} & \cellcolor[HTML]{FAFAFA}${n_l'=1}$ \\ \hline
\multicolumn{3}{|c|}{\cellcolor[HTML]{FAFAFA}Backbone}                                                                                                                                                             & \multicolumn{4}{c|}{\cellcolor[HTML]{FAFAFA}MobileNet-V2~\cite{sandler2018mobilenetv2}}                                                                                                                                                                 & \multicolumn{4}{c|}{\cellcolor[HTML]{FAFAFA}ResNet-50~\cite{resnet}}                                                                                                                                                                                    & \multicolumn{4}{c|}{\cellcolor[HTML]{FAFAFA}Swin-Base~\cite{swin}}                                                                                                                                                                                      \\ \hline
\multicolumn{1}{|l|}{\cellcolor[HTML]{FAFAFA}}                                                               & \multicolumn{1}{c|}{\cellcolor[HTML]{FAFAFA}}                                             & 18 Val  & \multicolumn{1}{c|}{\cellcolor[HTML]{F9FFF9}{83.8}}                                     & \multicolumn{1}{c|}{\cellcolor[HTML]{F9FBFF}\textbf{84.4}} & \multicolumn{1}{c|}{\cellcolor[HTML]{F9FBFF}83.7}       & 80.6                               & \multicolumn{1}{c|}{\cellcolor[HTML]{F9FFF9}84.1}                                           & \multicolumn{1}{c|}{\cellcolor[HTML]{F9FBFF}\textbf{85.0}} & \multicolumn{1}{c|}{\cellcolor[HTML]{F9FBFF}{84.5}} & 81.6                               & \multicolumn{1}{c|}{\cellcolor[HTML]{F9FFF9}84.5}                                           & \multicolumn{1}{c|}{\cellcolor[HTML]{F9FBFF}\textbf{84.9}} & \multicolumn{1}{c|}{\cellcolor[HTML]{F9FBFF}{84.7}} & 82.6                               \\ \cline{3-15} 
\multicolumn{1}{|l|}{\cellcolor[HTML]{FAFAFA}}                                                               & \multicolumn{1}{c|}{\multirow{-2}{*}{\cellcolor[HTML]{FAFAFA}$\mathcal{J}\&\mathcal{F}$}} & 19 Val  & \multicolumn{1}{c|}{\cellcolor[HTML]{F9FFF9}{83.7}}                                     & \multicolumn{1}{c|}{\cellcolor[HTML]{F9FBFF}\textbf{84.4}} & \multicolumn{1}{c|}{\cellcolor[HTML]{F9FBFF}83.6}       & 80.5                               & \multicolumn{1}{c|}{\cellcolor[HTML]{F9FFF9}84.1}                                           & \multicolumn{1}{c|}{\cellcolor[HTML]{F9FBFF}\textbf{84.9}} & \multicolumn{1}{c|}{\cellcolor[HTML]{F9FBFF}{84.3}} & 81.5                               & \multicolumn{1}{c|}{\cellcolor[HTML]{F9FFF9}84.5}                                           & \multicolumn{1}{c|}{\cellcolor[HTML]{F9FBFF}\textbf{85.0}} & \multicolumn{1}{c|}{\cellcolor[HTML]{F9FBFF}{84.8}} & 82.7                               \\ \cline{2-15} 
\multicolumn{1}{|l|}{\multirow{-3}{*}{\cellcolor[HTML]{FAFAFA}\makecell{YouTube-VOS\\\cite{youtubevos}}}}    & \multicolumn{2}{c|}{\cellcolor[HTML]{FAFAFA}FPS}                                                    & \multicolumn{1}{c|}{\cellcolor[HTML]{F9FFF9}16.0}                                           & \multicolumn{1}{c|}{\cellcolor[HTML]{F9FBFF}15.9}          & \multicolumn{1}{c|}{\cellcolor[HTML]{F9FBFF}{21.9}} & \textbf{34.4}                      & \multicolumn{1}{c|}{\cellcolor[HTML]{F9FFF9}14.9}                                           & \multicolumn{1}{c|}{\cellcolor[HTML]{F9FBFF}14.9}          & \multicolumn{1}{c|}{\cellcolor[HTML]{F9FBFF}{20.2}} & \textbf{30.9}                      & \multicolumn{1}{c|}{\cellcolor[HTML]{F9FFF9}9.3}                                            & \multicolumn{1}{c|}{\cellcolor[HTML]{F9FBFF}9.3}           & \multicolumn{1}{c|}{\cellcolor[HTML]{F9FBFF}{11.0}} & \textbf{13.5}                      \\ \hline
\multicolumn{1}{|c|}{\cellcolor[HTML]{FAFAFA}}                                                               & \multicolumn{1}{c|}{\cellcolor[HTML]{FAFAFA}}                                             & 17 Val  & \multicolumn{1}{c|}{\cellcolor[HTML]{F9FFF9}{83.8}}                                     & \multicolumn{1}{c|}{\cellcolor[HTML]{F9FBFF}\textbf{84.2}} & \multicolumn{1}{c|}{\cellcolor[HTML]{F9FBFF}83.3}       & 81.4                               & \multicolumn{1}{c|}{\cellcolor[HTML]{F9FFF9}84.9}                                           & \multicolumn{1}{c|}{\cellcolor[HTML]{F9FBFF}\textbf{85.6}} & \multicolumn{1}{c|}{\cellcolor[HTML]{F9FBFF}{85.3}} & 83.7                               & \multicolumn{1}{c|}{\cellcolor[HTML]{F9FFF9}{85.4}}                                     & \multicolumn{1}{c|}{\cellcolor[HTML]{F9FBFF}\textbf{85.5}} & \multicolumn{1}{c|}{\cellcolor[HTML]{F9FBFF}85.3}       & 83.0                               \\ \cline{3-15} 
\multicolumn{1}{|c|}{\cellcolor[HTML]{FAFAFA}}                                                               & \multicolumn{1}{c|}{\cellcolor[HTML]{FAFAFA}}                                             & 17 Test & \multicolumn{1}{c|}{\cellcolor[HTML]{F9FFF9}{78.3}}                                     & \multicolumn{1}{c|}{\cellcolor[HTML]{F9FBFF}\textbf{79.5}} & \multicolumn{1}{c|}{\cellcolor[HTML]{F9FBFF}78.1}       & 72.7                               & \multicolumn{1}{c|}{\cellcolor[HTML]{F9FFF9}{79.6}}                                     & \multicolumn{1}{c|}{\cellcolor[HTML]{F9FBFF}\textbf{79.9}} & \multicolumn{1}{c|}{\cellcolor[HTML]{F9FBFF}78.1}       & 71.2                               & \multicolumn{1}{c|}{\cellcolor[HTML]{F9FFF9}81.2}                                           & \multicolumn{1}{c|}{\cellcolor[HTML]{F9FBFF}\textbf{82.7}} & \multicolumn{1}{c|}{\cellcolor[HTML]{F9FBFF}{81.7}} & 78.6                               \\ \cline{3-15} 
\multicolumn{1}{|c|}{\cellcolor[HTML]{FAFAFA}}                                                               & \multicolumn{1}{c|}{\multirow{-3}{*}{\cellcolor[HTML]{FAFAFA}$\mathcal{J}\&\mathcal{F}$}} & 16 Val  & \multicolumn{1}{c|}{\cellcolor[HTML]{F9FFF9}90.4}                                           & \multicolumn{1}{c|}{\cellcolor[HTML]{F9FBFF}\textbf{91.6}} & \multicolumn{1}{c|}{\cellcolor[HTML]{F9FBFF}{90.9}} & 90.5                               & \multicolumn{1}{c|}{\cellcolor[HTML]{F9FFF9}91.1}                                           & \multicolumn{1}{c|}{\cellcolor[HTML]{F9FBFF}\textbf{92.1}} & \multicolumn{1}{c|}{\cellcolor[HTML]{F9FBFF}{92.0}} & 90.3                               & \multicolumn{1}{c|}{\cellcolor[HTML]{F9FFF9}92.0}                                           & \multicolumn{1}{c|}{\cellcolor[HTML]{F9FBFF}\textbf{92.4}} & \multicolumn{1}{c|}{\cellcolor[HTML]{F9FBFF}{92.2}} & 92.1                               \\ \cline{2-15} 
\multicolumn{1}{|c|}{\multirow{-4}{*}{\cellcolor[HTML]{FAFAFA}\makecell{DAVIS\\\cite{davis2017,davis2016}}}} & \multicolumn{2}{c|}{\cellcolor[HTML]{FAFAFA}FPS}                                                    & \multicolumn{1}{c|}{\cellcolor[HTML]{F9FFF9}18.7}                                           & \multicolumn{1}{c|}{\cellcolor[HTML]{F9FBFF}18.2}          & \multicolumn{1}{c|}{\cellcolor[HTML]{F9FBFF}{24.8}} & \textbf{38.6}                      & \multicolumn{1}{c|}{\cellcolor[HTML]{F9FFF9}18.0}                                           & \multicolumn{1}{c|}{\cellcolor[HTML]{F9FBFF}17.5}          & \multicolumn{1}{c|}{\cellcolor[HTML]{F9FBFF}{24.3}} & \textbf{37.4}                      & \multicolumn{1}{c|}{\cellcolor[HTML]{F9FFF9}12.1}                                           & \multicolumn{1}{c|}{\cellcolor[HTML]{F9FBFF}12.0}          & \multicolumn{1}{c|}{\cellcolor[HTML]{F9FBFF}{14.4}} & \textbf{17.9}                      \\ \hline
\end{tabular}
\end{center}
\end{table*}
\begin{table*}
\begin{center}
\vspace{-2.0mm}
\caption{\textbf{Cross-dataset evaluation (\S\ref{sec:compare}) on our VOSW benchmark with dense objects}. All the methods are fairly trained on YouTube-VOS~\cite{youtubevos} and DAVIS~\cite{davis2017} and evaluated on VOSW.}\label{tab:vosw}

\renewcommand\arraystretch{1.4}
\setlength{\tabcolsep}{7pt}
\begin{tabular}{|
>{\columncolor[HTML]{FAFAFA}}c |c|c
>{\columncolor[HTML]{FAFAFA}}c 
>{\columncolor[HTML]{FAFAFA}}c 
>{\columncolor[HTML]{FAFAFA}}c 
>{\columncolor[HTML]{FAFAFA}}c |
>{\columncolor[HTML]{FAFAFA}}c 
>{\columncolor[HTML]{FAFAFA}}c 
>{\columncolor[HTML]{FAFAFA}}c 
>{\columncolor[HTML]{FAFAFA}}c |}
\hline
\cellcolor[HTML]{FAFAFA}                         & \cellcolor[HTML]{FAFAFA}                                                 & \multicolumn{1}{c|}{\cellcolor[HTML]{FAFAFA}}                                                        & \multicolumn{1}{c|}{\cellcolor[HTML]{FAFAFA}}                                               & \multicolumn{3}{c|}{\cellcolor[HTML]{FAFAFA}AOST}                                                                                                   & \multicolumn{1}{c|}{\cellcolor[HTML]{FAFAFA}}                                               & \multicolumn{3}{c|}{\cellcolor[HTML]{FAFAFA}AOST}                                                                                                   \\ \cline{5-7} \cline{9-11} 
\multirow{-2}{*}{\cellcolor[HTML]{FAFAFA}Method} & \multirow{-2}{*}{\cellcolor[HTML]{FAFAFA}\makecell{CFBI+\cite{cfbip}}} & \multicolumn{1}{c|}{\multirow{-2}{*}{\cellcolor[HTML]{FAFAFA}\makecell{STCN\cite{cheng2021stcn}}}} & \multicolumn{1}{c|}{\multirow{-2}{*}{\cellcolor[HTML]{FAFAFA}\makecell{AOT-L\cite{aot}}}} & \multicolumn{1}{c|}{\cellcolor[HTML]{FAFAFA}${n_l'=3}$}    & \multicolumn{1}{c|}{\cellcolor[HTML]{FAFAFA}${n_l'=2}$} & ${n_l'=1}$                   & \multicolumn{1}{c|}{\multirow{-2}{*}{\cellcolor[HTML]{FAFAFA}\makecell{AOT-L\cite{aot}}}} & \multicolumn{1}{c|}{\cellcolor[HTML]{FAFAFA}${n_l'=3}$}    & \multicolumn{1}{c|}{\cellcolor[HTML]{FAFAFA}${n_l'=2}$} & ${n_l'=1}$                   \\ \hline
Backbone                                         & \cellcolor[HTML]{FAFAFA}ResNet-101~\cite{resnet}                         & \multicolumn{5}{c|}{\cellcolor[HTML]{FAFAFA}ResNet-50~\cite{resnet}}                                                                                                                                                                                                                                                                                     & \multicolumn{4}{c|}{\cellcolor[HTML]{FAFAFA}Swin-Base~\cite{swin}}                                                                                                                                                                                \\ \hline
$\mathcal{J}$                                    & 73.3                                                                     & \multicolumn{1}{c|}{74.1}                                                                            & \multicolumn{1}{c|}{\cellcolor[HTML]{F9FFF9}{75.8}}                                     & \multicolumn{1}{c|}{\cellcolor[HTML]{F9FBFF}\textbf{76.5}} & \multicolumn{1}{c|}{\cellcolor[HTML]{F9FBFF}75.7}       & \cellcolor[HTML]{F9FBFF}73.7 & \multicolumn{1}{c|}{\cellcolor[HTML]{F9FFF9}{75.7}}                                     & \multicolumn{1}{c|}{\cellcolor[HTML]{F9FBFF}\textbf{76.1}} & \multicolumn{1}{c|}{\cellcolor[HTML]{F9FBFF}75.3}       & \cellcolor[HTML]{F9FBFF}73.4 \\ \hline
$\mathcal{F}$                                    & 71.5                                                                     & \multicolumn{1}{c|}{72.0}                                                                            & \multicolumn{1}{c|}{\cellcolor[HTML]{F9FFF9}{73.7}}                                     & \multicolumn{1}{c|}{\cellcolor[HTML]{F9FBFF}\textbf{74.5}} & \multicolumn{1}{c|}{\cellcolor[HTML]{F9FBFF}73.6}       & \cellcolor[HTML]{F9FBFF}71.4 & \multicolumn{1}{c|}{\cellcolor[HTML]{F9FFF9}{73.4}}                                     & \multicolumn{1}{c|}{\cellcolor[HTML]{F9FBFF}\textbf{73.9}} & \multicolumn{1}{c|}{\cellcolor[HTML]{F9FBFF}73.0}       & \cellcolor[HTML]{F9FBFF}70.8 \\ \hline
$\mathcal{J}\&\mathcal{F}$                       & 72.4                                                                     & \multicolumn{1}{c|}{73.1}                                                                            & \multicolumn{1}{c|}{\cellcolor[HTML]{F9FFF9}{74.8}}                                     & \multicolumn{1}{c|}{\cellcolor[HTML]{F9FBFF}\textbf{75.5}} & \multicolumn{1}{c|}{\cellcolor[HTML]{F9FBFF}74.7}       & \cellcolor[HTML]{F9FBFF}72.6 & \multicolumn{1}{c|}{\cellcolor[HTML]{F9FFF9}{74.6}}                                     & \multicolumn{1}{c|}{\cellcolor[HTML]{F9FBFF}\textbf{75.0}} & \multicolumn{1}{c|}{\cellcolor[HTML]{F9FBFF}74.2}       & \cellcolor[HTML]{F9FBFF}72.1 \\ \hline
\end{tabular}
\end{center}
\vspace{-3mm}
\end{table*}

\noindent \textbf{Cross-dataset Evaluation on VOSW}. To further validate our approaches' generalization ability in scenarios with dense objects, we compare AOST/AOT with state-of-the-art template-based method, CFBI+~\cite{cfbip} and attention-based method, STCN~\cite{cheng2021stcn} on our challenging VOSW benchmark (\S\ref{sec:VOSW}). As shown in Table~\ref{tab:vosw}, our methods based on identification mechanisms are proficient in tracking multiple objects, resulting in superior performance. For example, with the same ResNet-50 backbone, AOT significantly outperforms STCN (\textbf{74.8\%} \vs 73.1). By introducing S-LSTT, AOST$^{n_l'=3}$ further boost the performance of AOT to \textbf{75.5\%}. Notably, the results indicate that the utilization of a large backbone (Swin-Base~\cite{swin}) may lead to a mild performance decrement (\eg, from 74.8\% to 74.6\%) during cross-dataset evaluation. The decrement could be attributed to the overfitting of training datasets due to the utilization of excessive parameters.

\noindent \textbf{Scalability of AOST}. To validate AOST's scalability, we conduct a series of experiments in Table~\ref{tab:aot_aost_compare} and Table~\ref{tab:vosw}. The results show that AOST has excellent online flexibility and can be adapted between state-of-the-art performance and real-time speed. Notably, AOST$^{n_l=3}$ \textbf{consistently outperforms} AOT on all six benchmarks with three different backbones. Moreover, AOST$^{n_l=2}$ is \textbf{consistently more efficient} than AOT and is comparable with (or even surpasses) AOT in most cases. By adjusting $n_l'$ to 1 during inference, the speed of AOST can be further boosted to \textbf{\textgreater35FPS}, which is about 2$\times$ faster than the AOT counterpart (AOT-L).

\noindent \textbf{Single-object Evaluation on DAVIS 2016}~\cite{davis2016}. \new{Although our approaches aim at improving multi-object VOS, we also achieve promising performance on the single-object DAVIS 2016 Val~\cite{davis2016} (\eg, R50-AOST/SwinB-AOST \textbf{92.1\%/92.4\%}). Under single-object scenarios, the multi-object superiority of AOST is limited, but R50-AOST$^{n_l'=2}$ (shown in Table~\ref{tab:aot_aost_compare}) still maintains a comparable efficiency compared to STCN (\textbf{24.3} \vs 27.2FPS) and performs better (\textbf{92.0\%} \vs 91.6\%). Furthermore, our AOST$^{n_l'=1}$ achieves comparable performance with recent RPCM~\cite{rpcm} (\textbf{90.5\%} \vs 90.6\%) while running about \textbf{6.7$\times$} faster (\textbf{38.6} \vs 5.8FPS).}


\noindent\textbf{Test-time Augmentations.} Table~\ref{tab:aug} shows we can further boost AOST's performance. First, using All-Frame (30FPS) videos (which is also a common strategy used by attention-based methods as shown in Table~\ref{tab:youtubevos}) on YouTube-VOS always leads to better performance. In addition, we can further boost AOST's performance by utilizing multi-scale and flipping augmentations. At last, our SwinB-AOST achieves state-of-the-art performance on commonly-used benchmarks, YouTube-VOS 2018/2019 Val (\textbf{86.5\%/86.5\%}), DAVIS 2017 Val/Test (\textbf{87.0\%/84.7\%}), and DAVIS 2016 (\textbf{93.0\%}).

\noindent\textbf{Qualitative Results}. Fig.~\ref{fig:comparisons} shows that our approaches perform better than templated-based CFBI~\cite{cfbi,cfbip} when segmenting multiple highly similar objects on YouTube-VOS and DAVIS. By introducing S-LSTT, AOST is more robust than AOT and produces finer results. On our challenging VOSW with dense objects, our approaches significantly outperform STCN~\cite{cheng2021stcn}, a state-of-the-art attention-based method. As shown in Fig.~\ref{fig:vosw_comparisons}, STCN fails to track and segment numerous moving objects, while our approaches produce sharp segmentation results with the identification mechanism. Also, our AOST with S-LSTT outperforms AOT when tracking highly similar objects in complex scenarios. Nevertheless, current methods are difficult to track tiny objects (lacking reference information) and highly deformed objects (with severe inconsistency between the current object and the reference object). We believe future studies with specific designs promise to address these issues.



\subsection{Ablation Study}\label{sec:ablation}
In this section, we analyze the main components and hyper-parameters of our approaches in Table~\ref{tab:ablation} and~\ref{tab:ablation_aost}.

\subsubsection{ID Mechanism and Long Short-term Attention}\label{sec:ablation_aot}

\noindent\textbf{Identity Number:} The number of the identification vectors, $n_{id}$, must exceed the number of objects. Thus, we set the default value of $n_{id}$ to 10 to align with the maximum object count in the training splits~\cite{youtubevos,davis2017}. Table~\ref{tab:id_number} illustrates that $n_{id}$ larger than 10 leads to performance degradation. 
One possible explanation for this is that none of the training videos has enough objects that can be assigned \textgreater10 identities. Therefore, the network fails to learn all identities simultaneously.
Furthermore, to establish that the multi-object association indeed benefits our approaches, we impose $n_{id}=1$ and use the post-ensemble strategy for inference. As expected, this results in a decrease in performance (from 80.3\% to 78.7\%).

\begin{table}[t!]
	\centering
	\caption{\textbf{Quantitative results (\S\ref{sec:compare}) on DAVIS 2017 Val}~\cite{davis2017}. 
        $^{\ddag}$: timing extrapolated from single-object speed assuming linear scaling in the number of objects. 
	}\label{tab:davis}
	\vspace{-3.0mm}
\centering
\setlength{\tabcolsep}{11pt}
\renewcommand\arraystretch{1.05}
\begin{tabular}{rcccc}
\bottomrule[1pt]\rowcolor[HTML]{FAFAFA}
Method                                  & $\mathcal{J}\&\mathcal{F}$ & $\mathcal{J}$ & $\mathcal{F}$ & FPS                                                                                       \\
\toprule[0.8pt]
STM\hspace{0.68em}\hspace{0.45em}\pub{ICCV19}\hspace{0.77em}\hspace{0.48em}\cite{spacetime}         & 81.8                       & 79.2          & 84.3          & 3.1$^\ddag$   \\
EGMN\hspace{0.68em}~\pub{ECCV20}\hspace{0.77em}\hspace{0.48em}\cite{EGMN}             & 82.8                       & 80.2          & 85.2          & 5.0           \\
LWL\hspace{0.68em}~\pub{ECCV20}\hspace{0.77em}\cite{LWLVOS}            & 81.6                       & 79.1          & 84.1          & 2.5$^\ddag$   \\
KMN\hspace{0.68em}~\pub{ECCV20}\hspace{0.77em}\hspace{0.48em}\cite{KMN}               & 82.8                       & 80.0          & 85.6          & 4.2$^\ddag$   \\
CFBI\hspace{0.68em}~\pub{ECCV20}\hspace{0.77em}\hspace{0.48em}\cite{cfbi}             & 81.9                       & 79.3          & 84.5          & 5.9           \\
SST\hspace{0.68em}~\pub{CVPR21}~\cite{sstvos}            & 82.5                       & 79.9          & 85.1          & -             \\
Joint\hspace{0.68em}~\pub{CVPR21}\hspace{0.77em}\cite{mao2021joint}    & 83.5                       & 80.8          & 86.2          & 4.0           \\
CFBI+\hspace{0.68em}\hspace{0.35em}\pub{PAMI21}\hspace{0.77em}\hspace{0.48em}\cite{cfbip}          & 82.9                       & 80.1          & 85.7          & 5.6           \\
HMMN\hspace{0.78em}~\pub{ICCV21}~\cite{hmmn}             & 84.7                       & 81.9          & 87.5          & 5.0$^\ddag$   \\
STCN~\pub{NeurIPS21}\hspace{0.77em}\hspace{0.48em}\cite{cheng2021stcn} & 85.4                       & 82.2          & \underline{88.6}    & 24.7  \\
RPCM\hspace{0.68em}~\pub{AAAI22}\hspace{0.77em}\cite{rpcm}             & 83.7                       & 81.3          & 86.0          & 2.9$^\ddag$   \\
GSFM\hspace{0.68em}~\pub{ECCV22}\hspace{0.77em}\cite{liu2022learning}        & 86.2                       & \textbf{83.1}          & \underline{89.3}          & 8.9           \\
\hline
\hline
\rowcolor[HTML]{F9FFF9}
\rowcolor[HTML]{F9FFF9}
AOT-S~\pub{NeurIPS21}\hspace{0.77em}\cite{aot}                                      & 81.3                       & 78.7          & 83.9          & \textbf{{40.0}}  \\
\rowcolor[HTML]{F9FFF9}
AOT-B~\pub{NeurIPS21}\hspace{0.77em}\cite{aot}                                      & 82.5                       & 79.7          & 85.2          & \underline{29.6}          \\
\rowcolor[HTML]{F9FFF9}
AOT-L~\pub{NeurIPS21}\hspace{0.77em}\cite{aot}                                      & {83.8}                     & {81.1}        & {86.4}        & 18.7          \\
\hdashline
\rowcolor[HTML]{F9FBFF}
AOST$^{n_l'=3}$\hspace{0.5em}                          & 84.2                       & 81.2          & 87.2          & {18.2}        \\
\rowcolor[HTML]{F9FBFF}
R50-AOST$^{n_l'=3}$\hspace{0.5em}                       & \underline{85.6}                 & \underline{82.6}    & 88.5          & {17.5}        \\
\rowcolor[HTML]{F9FBFF}
R50-AOST$^{n_l'=3}$$^\dagger$            & {85.3}                     & {82.3}        & {88.2}        & 17.5          \\
\rowcolor[HTML]{F9FBFF}
SwinB-AOST$^{n_l'=3}$\hspace{0.5em}                     & 85.5                       & 82.4          & {88.6}    & {12.0}        \\
\rowcolor[HTML]{F9FBFF}
SwinB-AOST$^{n_l'=3}$$^\dagger$          & \textbf{86.3}              & \textbf{83.1} & \textbf{89.4} & 12.0          \\
\toprule[1pt]
\end{tabular}\vspace{-2mm}
\end{table}

\begin{table}[!t]
\begin{center}

\caption{\textbf{Performance ($\mathcal{J}\&\mathcal{F}$) with test-time augmentations on YouTube-VOS~\cite{youtubevos} and DAVIS~\cite{davis2016,davis2017} (\S\ref{sec:compare})}. MS: using multi-scale and flipping augmentations.}\label{tab:aug}\vspace{-2mm}
\renewcommand\arraystretch{1.2}
\setlength{\tabcolsep}{3pt}
\begin{tabular}{|l|c|ccc|ccc|}
\hline
\rowcolor[HTML]{FAFAFA} 
\cellcolor[HTML]{FAFAFA}                          & \cellcolor[HTML]{FAFAFA}                     & \multicolumn{3}{c|}{\cellcolor[HTML]{FAFAFA}YouTube-VOS~\cite{youtubevos}}                                            & \multicolumn{3}{c|}{\cellcolor[HTML]{FAFAFA}DAVIS~\cite{davis2016,davis2017}}                                                             \\ \cline{3-8} 
\rowcolor[HTML]{FAFAFA} 
\multirow{-2}{*}{\cellcolor[HTML]{FAFAFA}Method}  & \multirow{-2}{*}{\cellcolor[HTML]{FAFAFA}MS} & \multicolumn{1}{c|}{\cellcolor[HTML]{FAFAFA}AF} & \multicolumn{1}{c|}{\cellcolor[HTML]{FAFAFA}18 Val} & 19 Val        & \multicolumn{1}{c|}{\cellcolor[HTML]{FAFAFA}17 Val} & \multicolumn{1}{c|}{\cellcolor[HTML]{FAFAFA}17 Test} & 16 Val                       \\ \hline
                                                  &                                              & \multicolumn{1}{c|}{}                           & \multicolumn{1}{c|}{84.8}                           & 85.0          & \multicolumn{1}{c|}{}                               & \multicolumn{1}{c|}{}                                &                              \\ \cline{2-5}
                                                  &                                              & \multicolumn{1}{c|}{\checkmark}                 & \multicolumn{1}{c|}{85.4}                           & 85.3          & \multicolumn{1}{c|}{\multirow{-2}{*}{85.3}}         & \multicolumn{1}{c|}{\multirow{-2}{*}{80.3}}          & \multirow{-2}{*}{91.4}       \\ \cline{2-8} 
\multirow{-3}{*}{SwinB-AOST$^{n_l'=3}$}           & \checkmark                                   & \multicolumn{1}{c|}{\checkmark}                 & \multicolumn{1}{c|}{\underline{86.2}}                     & \underline{86.3}    & \multicolumn{1}{c|}{\underline{86.7}}                     & \multicolumn{1}{c|}{\textbf{84.7}}                   & \textbf{93.0}                \\ \hline
                                                  &                                              & \multicolumn{1}{c|}{}                           & \multicolumn{1}{c|}{85.1}                           & 85.2          & \multicolumn{1}{c|}{}                               & \multicolumn{1}{c|}{}                                &                              \\ \cline{2-5}
                                                  &                                              & \multicolumn{1}{c|}{\checkmark}                 & \multicolumn{1}{c|}{85.8}                           & 86.0          & \multicolumn{1}{c|}{\multirow{-2}{*}{86.3}}         & \multicolumn{1}{c|}{\multirow{-2}{*}{81.9}}          & \multirow{-2}{*}{\underline{92.4}} \\ \cline{2-8} 
\multirow{-3}{*}{SwinB-AOST$^{n_l'=3}$$^\dagger$} & \checkmark                                   & \multicolumn{1}{c|}{\checkmark}                 & \multicolumn{1}{c|}{\textbf{86.5}}                  & \textbf{86.5} & \multicolumn{1}{c|}{\textbf{87.0}}                  & \multicolumn{1}{c|}{\underline{84.5}}                      & \textbf{93.0}                \\ \hline
\end{tabular}
\end{center}
\vspace{-3.0mm}
\end{table}


\noindent\textbf{Local Window Size:} \new{Table~\ref{tab:window_size} indicates that a suitable local window size, denoted by $\lambda$, leads to a better performance-efficiency balance. Specifically, the absence of local attention ($\lambda=0$) causes a significant decline in performance (from 80.3\% to 74.3\%), further highlighting the crucial role of local attention. Further increase $\lambda$ from 15 to 19 only slightly increases the performance (from 80.3\% to 80.4\%), while 27\% more memory (0.58G \vs 0.73G) will be taken.}


\noindent\textbf{Local Frame Number:}$_{\!}$ In$_{\!}$ Table$_{\!}$~\ref{tab:local_frame},$_{\!}$ we$_{\!}$ endeavor$_{\!}$ to$_{\!}$ integrate$_{\!}$ more$_{\!}$ previous$_{\!}$ frames$_{\!}$ into$_{\!}$ local$_{\!}$ attention$_{\!}$ but$_{\!}$ observe$_{\!}$ that$_{\!}$ leveraging$_{\!}$ only$_{\!}$ $t-1$$_{\!}$ frame$_{\!}$ (80.3\%)$_{\!}$ yielded$_{\!}$ better performance compared to utilizing$_{\!}$ 2/3$_{\!}$ frames$_{\!}$ (80.0\%/79.1\%)$_{\!}$. This$_{\!}$ outcome$_{\!}$ could$_{\!}$ stem$_{\!}$ from$_{\!}$ the$_{\!}$ fact$_{\!}$ that$_{\!}$ the$_{\!}$ motion$_{\!}$ between$_{\!}$ frames becomes more pronounced with a longer temporal interval. As such, a local matching that employs an earlier previous frame runs a higher risk of introducing errors.



\noindent\textbf{Block Number:}$_{\!}$ In$_{\!}$ Table$_{\!}$~\ref{tab:lstt_number},$_{\!}$ AOT's$_{\!}$ performance$_{\!}$ is$_{\!}$ improved$_{\!}$ by$_{\!}$ adding$_{\!}$ more$_{\!}$ LSTT$_{\!}$ blcoks.$_{\!}$ Remarkably,$_{\!}$ the$_{\!}$ AOT$_{\!}$ using$_{\!}$ only$_{\!}$ one$_{\!}$ block$_{\!}$ attains$_{\!}$ a$_{\!}$ rapid$_{\!}$ real-time$_{\!}$ speed$_{\!}$ (41.0FPS)$_{\!}$, albeit$_{\!}$ its$_{\!}$ performance$_{\!}$ is$_{\!}$ inferior$_{\!}$ by$_{\!}$ -2.4\%$_{\!}$ in$_{\!}$ contrast$_{\!}$ to$_{\!}$ two$_{\!}$ blocks$_{\!}$ (80.3\%).$_{\!}$ By$_{\!}$ manipulating$_{\!}$ the$_{\!}$ number$_{\!}$ of$_{\!}$ blocks,$_{\!}$ we can control the balance between accuracy and speed.

\begin{table}[t!]
	\centering
	\caption{\textbf{Quantitative results (\S\ref{sec:compare}) on DAVIS 2017 Test}~\cite{davis2017}. $^{*}$: using 600p instead of 480p videos in inference.
	}\label{tab:davis_test}
	\vspace{-3.0mm}
\centering
\setlength{\tabcolsep}{11pt}
\renewcommand\arraystretch{1.05}
\begin{tabular}{rcccc}
\bottomrule[1pt]\rowcolor[HTML]{FAFAFA}
Method                                  & $\mathcal{J}\&\mathcal{F}$ & $\mathcal{J}$ & $\mathcal{F}$ & FPS           \\
\toprule[0.8pt]
STM$^{*}$\hspace{0.18em}\hspace{0.45em}\pub{ICCV19}\hspace{0.48em}\hspace{0.77em}\cite{spacetime}   & 72.2                       & 69.3          & 75.2          & -             \\
CFBI\hspace{0.68em}~\pub{ECCV20}\hspace{0.77em}\hspace{0.48em}\cite{cfbi}             & 75.0                       & 71.4          & 78.7          & 5.3           \\
CFBI$^{*}$\hspace{0.18em}~\pub{ECCV20}\hspace{0.77em}\hspace{0.48em}\cite{cfbi}       & 76.6                       & 73.0          & 80.1          & 2.9           \\
KMN$^{*}$\hspace{0.18em}~\pub{ECCV20}\hspace{0.77em}\hspace{0.48em}\cite{KMN}         & 77.2                       & 74.1          & 80.3          & 2.8$^\ddag$   \\
STCN~\pub{NeurIPS21}\hspace{0.77em}\hspace{0.48em}\cite{cheng2021stcn} & 76.1                       & 72.7          & 79.6          & 19.5  \\
CFBI+$^{*}$\hspace{0.28em}\hspace{0.35em}\pub{PAMI21}\hspace{0.77em}\hspace{0.48em}\cite{cfbip}    & 78.0                       & 74.4          & 81.6          & 3.4           \\
HMMN\hspace{0.78em}~\pub{ICCV21}~\cite{hmmn}             & 78.6                       & 74.7          & 82.5          & 3.4$^\ddag$   \\
RPCM\hspace{0.68em}~\pub{AAAI22}\hspace{0.77em}\cite{rpcm}             & 79.2                       & 75.8          & 82.6          & 1.9$^\ddag$   \\
GSFM\hspace{0.68em}~\pub{ECCV22}\hspace{0.77em}\cite{liu2022learning}             & 77.5                       & 74.0          & 80.9          & 8.9   \\
\hline
\hline
\rowcolor[HTML]{F9FFF9}
\rowcolor[HTML]{F9FFF9}
AOT-S~\pub{NeurIPS21}\hspace{0.77em}\cite{aot}                                      & 73.9                       & 70.3          & 77.5          & \textbf{40.0}    \\
\rowcolor[HTML]{F9FFF9}
AOT-B~\pub{NeurIPS21}\hspace{0.77em}\cite{aot}                                      & 75.5                       & 71.6          & 79.3          & \underline{29.6}          \\
\rowcolor[HTML]{F9FFF9}
AOT-L~\pub{NeurIPS21}\hspace{0.77em}\cite{aot}                                      & 78.3                       & 74.3          & 82.3          & 18.7          \\
\hdashline
\rowcolor[HTML]{F9FBFF}
AOST$^{n_l'=3}$\hspace{0.5em}                          & 79.5                       & 75.6          & 83.3          & {18.2}        \\
\rowcolor[HTML]{F9FBFF}
R50-AOST$^{n_l'=3}$\hspace{0.5em}                      & 79.9                       & 76.2          & 83.6          & {17.5}        \\
\rowcolor[HTML]{F9FBFF}
R50-AOST-L$^{n_l'=3}$$^\dagger$          & {80.3}                     & {76.7}        & {83.9}        & 17.5          \\
\rowcolor[HTML]{F9FBFF}
SwinB-AOST$^{n_l'=3}$\hspace{0.5em}                   & \textbf{82.7}              & \textbf{78.8} & \textbf{86.6} & {12.0}        \\
\rowcolor[HTML]{F9FBFF}
SwinB-AOST$^{n_l'=3}$$^\dagger$          & \underline{{81.9}}               & \underline{{78.2}}  & \underline{{85.6}}  & 12.0          \\
\toprule[1pt]
\end{tabular}\vspace{-2.0mm}
\end{table}

\begin{table}
\begin{center}

\caption{\textbf{Quantitative results (\S\ref{sec:compare}) on DAVIS 2016}~\cite{davis2016}.}\label{tab:davis2016}
\vspace{-2.0mm}
\renewcommand\arraystretch{1.05}
\setlength{\tabcolsep}{10pt}
\begin{tabular}{rcccc}
\bottomrule[1.pt]\rowcolor[HTML]{FAFAFA}
Method                                  & $\mathcal{J}\&\mathcal{F}$ & $\mathcal{J}$            & $\mathcal{F}$            & FPS                      \\
\hline
STM\hspace{0.68em}\hspace{0.45em}\pub{ICCV19}\hspace{0.77em}\hspace{0.48em}\cite{spacetime}         & 89.3                       & 88.7                     & 89.9                     & 6.3                      \\
CFBI\hspace{0.68em}~\pub{ECCV20}\hspace{0.77em}\hspace{0.48em}\cite{cfbi}             & 89.4                       & 88.3                     & 90.5                     & 6.3                      \\
CFBI+\hspace{0.68em}\hspace{0.35em}\pub{PAMI21}\hspace{0.77em}\hspace{0.48em}\cite{cfbip}          & 89.9                       & 88.7                     & 91.1                     & 5.9                      \\
HMMN\hspace{0.78em}~\pub{ICCV21}~\cite{hmmn}             & 90.8                       & {89.6}                   & {92.0}                   & 10.0                     \\
STCN~\pub{NeurIPS21}\hspace{0.77em}\hspace{0.48em}\cite{cheng2021stcn} & 91.6                       & \textbf{{90.8}}          & {92.5}                   & 27.2$^\star$             \\
RPCM\hspace{0.68em}~\pub{AAAI22}\hspace{0.77em}\cite{rpcm}             & 90.6                       & 87.1                     & 94.0                     & 5.8                      \\
GSFM\hspace{0.68em}~\pub{ECCV22}\hspace{0.77em}\cite{liu2022learning}             & 91.4                       & 90.1          & 92.7          & 8.9   \\
\hline
\hline
\rowcolor[HTML]{F9FFF9}
\rowcolor[HTML]{F9FFF9}
AOT-S~\pub{NeurIPS21}\hspace{0.77em}\cite{aot}          & 89.4                       & 88.6                     & 90.2                     & \textbf{40.0}               \\
\rowcolor[HTML]{F9FFF9}
AOT-B~\pub{NeurIPS21}\hspace{0.77em}\cite{aot}          & 89.9                       & 88.7                     & 91.1                     & \underline{29.6}                     \\
\rowcolor[HTML]{F9FFF9}
AOT-L~\pub{NeurIPS21}\hspace{0.77em}\cite{aot}          & {90.4}                     & {89.6}                   & {91.1}                   & 18.7                     \\
\hdashline
\rowcolor[HTML]{F9FBFF}
AOST$^{n_l'=3}$\hspace{0.5em}                          & 91.6                       & 90.1                     & 93.0                     & {18.2}                   \\
\rowcolor[HTML]{F9FBFF}
R50-AOST$^{n_l'=3}$\hspace{0.5em}                       & \underline{92.1}                 & \underline{90.6}               & 93.6                     & {17.5}                   \\
\rowcolor[HTML]{F9FBFF}
R50-AOST$^{n_l'=3}$$^\dagger$          & {91.4}                     & {90.3}                   & {92.5}                   & 17.5                     \\
\rowcolor[HTML]{F9FBFF}
SwinB-AOST$^{n_l'=3}$\hspace{0.5em}                    & \textbf{92.4}              & 90.5                     & \textbf{94.2}            & {12.0}                   \\
\rowcolor[HTML]{F9FBFF}
SwinB-AOST$^{n_l'=3}$$^\dagger$        & \textbf{92.4}              & \underline{90.6}               & \underline{{94.1}}             & 12.0                     \\
\toprule[1.pt]
\end{tabular}
\end{center}
\vspace{-3.0mm}
\end{table}

\subsubsection{S-LSTT and Layer-wise ID-based Attention}\label{sec:ablation_aost}

\noindent\textbf{Scaling Weight:} To choose a suitable scaling weight $\alpha$, we train AOST with $\alpha$ ranging from 1 to 8, as shown in Table~\ref{tab:balance_weight}. When $\alpha\in[1,4]$, a larger $\alpha$ results in an enhanced performance of AOST with $n_l'=2/3$, but a declining performance of $n_l'=1$, since greater weights are assigned to the sub-AOST with deeper S-LSTT depth in the loss function (Eq.~\ref{eq:scalable_loss}). However, increasing $\alpha$ to 8 or larger value (\eg, setting $\alpha=\infty$ leads to performance degradation of R50-AOST$^{n_l'=3}$ in Table~\ref{tab:youtubevos}) decreases the training stability. As a result, we prefer to set $\alpha=2$ as the default setting.


\noindent\textbf{Scalable LSTT:} \new{Table~\ref{tab:lstt_type} shows the performances of AOST with varing S-LSTT types. Sharing parameters among different S-LSTT blocks, except for layer-wise identification and gating weights, reduces the total parameters of AOST from 8.8M to 5.8M, and the model still yields satisfactory results with a slight reduction in accuracy (80.3/82.8/83.6\% \vs 80.6/83.7/84.4\%). If we substitute S-LSTT with LSTT~\cite{aot} by removing layer-wise ID-based attention and scalable supervision, AOST would lose scalability, and the performance would drop from $84.4$ to $83.7$. Notably, sharing parameters among LSTT blocks leads to a performance reduction from $83.7$ to $82.8$. Based on the above comparisons, S-LSTT is more effective and flexible than LSTT by introducing layer-wise ID-based attention and scalable supervision, especially when sharing parameters among blocks.}

\begin{table*}[t!]
	\centering
	\caption{\textbf{Ablation study of identification mechanism and long short-term attention on YouTube-VOS~\cite{youtubevos} (\S\ref{sec:ablation_aot})}. We use AOT-S without pre-training on synthetic videos for simplicity. $\mathcal{J}_{S}$/$\mathcal{J}_{U}$: $\mathcal{J}$ on seen/unseen classes. 
 \new{Mem: memory usage (G) when processing a 480p video}.}\label{tab:ablation}
 \vspace{-2mm}
\renewcommand\arraystretch{1.2}
\begin{subtable}{.22\textwidth}
\center
\caption{Identity number}\label{tab:id_number}\vspace{-1mm}
\setlength{\tabcolsep}{5pt}
\begin{tabular}{|c|ccc|}
\hline
\rowcolor[HTML]{FAFAFA} 
$n_{id}$ & $\mathcal{J}\&\mathcal{F}$ & $\mathcal{J}_{S}$ & $\mathcal{J}_{U}$ \\ \hline
\textbf{10}  & \textbf{80.3}              & \textbf{80.6}     & \textbf{73.7}     \\ \hline
20  & 78.3                       & 79.4              & 70.8              \\
30  & 77.2                       & 78.5              & 70.2              \\ \hline
1   & \underline{78.7}                     & \underline{78.0}            & \underline{73.0}            \\ \hline
\end{tabular}
\end{subtable}
\begin{subtable}{.27\textwidth}
\center
\caption{Local window size}\label{tab:window_size}\vspace{-1mm}
\setlength{\tabcolsep}{5pt}
\begin{tabular}{|c|cccc|}
\hline
\rowcolor[HTML]{FAFAFA} 
$\lambda$   & $\mathcal{J}\&\mathcal{F}$ & $\mathcal{J}_{S}$ & $\mathcal{J}_{U}$ & \new{Mem} \\ \hline
\new{19}          & \new{\textbf{80.4}}                       & \new{\textbf{80.8}}            & \new{\underline{73.6}} & \new{0.73}             \\ \hline
\textbf{15} & \underline{80.3}              & \underline{80.6}     & \textbf{73.7} & \new{0.58}    \\ \hline
11          & 78.8                       & 79.5              & 71.9         &  \new{\underline{0.47}}    \\
0           & 74.3                       & 74.9              & 67.6       & \new{\textbf{0.37}}       \\ 
\hline
\end{tabular}
\end{subtable}
\begin{subtable}{.22\textwidth}
\center
\caption{Local frame number}\label{tab:local_frame}\vspace{-1mm}
\setlength{\tabcolsep}{5pt}
\begin{tabular}{|c|ccc|}
\hline
\rowcolor[HTML]{FCFCFC} 
${n_\mathrm{ST}}$        & $\mathcal{J}\&\mathcal{F}$ & $\mathcal{J}_{S}$ & $\mathcal{J}_{U}$ \\ \hline
\textbf{1} & \textbf{80.3}              & \textbf{80.6}     & \textbf{73.7}     \\ \hline
2          & 80.0                       & 79.8              & 73.7              \\
3          & 79.1                       & 80.0              & 72.2              \\
0          & 74.3                       & 74.9              & 67.6              \\ \hline
\end{tabular}
\end{subtable}\hspace{2mm}
\begin{subtable}{.24\textwidth}
\center
\caption{Block number}\label{tab:lstt_number}\vspace{-1mm}
\setlength{\tabcolsep}{5pt}
\begin{tabular}{|c|cccc|}
\hline
\rowcolor[HTML]{FCFCFC} 
$n_l$ & $\mathcal{J}\&\mathcal{F}$ & $\mathcal{J}_{S}$ & $\mathcal{J}_{U}$ & FPS           \\ \hline
1   & 77.9                       & 78.8              & 71.0              & \textbf{41.0} \\ \hline
2   & \underline{80.3}                     & \underline{80.6}            & \underline{73.7}            & \underline{27.1}          \\ \hline
3   & \textbf{80.9}              & \textbf{81.1}     & \textbf{74.0}     & 20.5          \\ \hline
\end{tabular}
\end{subtable}
\vspace{-1.0mm}
\end{table*}
\begin{table*}[t!]
	\centering
	\caption{\textbf{Ablation study of S-LSTT and layer-wise ID-based attention on YouTube-VOS~\cite{youtubevos} (\S\ref{sec:ablation_aost})}. The experiments are based on AOST, and the evaluation metric is $\mathcal{J}\&\mathcal{F}$ on YouTube-VOS~\cite{youtubevos}.}\label{tab:ablation_aost}
 \vspace{-4mm}
\center
\renewcommand\arraystretch{1.2}
\begin{subtable}{.23\textwidth}
\center
\caption{Scaling weight}\label{tab:balance_weight}\vspace{-1mm}
\setlength{\tabcolsep}{7pt}
\begin{tabular}{|c|ccc|}
\hline
\rowcolor[HTML]{FCFCFC} 
\cellcolor[HTML]{FCFCFC}                                            & \multicolumn{3}{c|}{\cellcolor[HTML]{FCFCFC}$n_l'$} \\ \cline{2-4} 
\rowcolor[HTML]{FCFCFC} 
\multirow{-2}{*}{\cellcolor[HTML]{FCFCFC}{$\alpha$}} & 1               & 2               & 3              \\ \hline
\textbf{2}                                                          & \underline{{80.6}}    & \underline{{83.7}}    & \underline{{84.4}}   \\ \hline
1                                                                   & \textbf{81.3}   & \textbf{83.9}   & 84.2           \\
4                                                                   & 79.8            & 83.5            & \textbf{84.5}  \\
8                                                                   & 79.5            & 83.3            & 84.1           \\ \hline
\end{tabular}
\end{subtable}
\begin{subtable}{.38\textwidth}
\center
\caption{Scalable LSTT}\label{tab:lstt_type}\vspace{-1mm}
\setlength{\tabcolsep}{5pt}
\begin{tabular}{|l|ccc|c|}
\hline
\rowcolor[HTML]{FCFCFC} 
\cellcolor[HTML]{FCFCFC}                                             & \multicolumn{3}{c|}{\cellcolor[HTML]{FCFCFC}$n_l'$} & \cellcolor[HTML]{FCFCFC}                        \\ \cline{2-4}
\rowcolor[HTML]{FCFCFC} 
\multirow{-2}{*}{\cellcolor[HTML]{FCFCFC}{LSTT Type}} & 1               & 2              & 3              & \multirow{-2}{*}{\cellcolor[HTML]{FCFCFC}Param} \\ \hline
S-LSTT                                                               & \textbf{80.6}   & \textbf{83.7}  & \textbf{84.4}  & 8.8                                             \\ \hline
Share S-LSTT layers                                                  & 80.3            & 82.8           & 83.6           & 5.8                                             \\
LSTT~\cite{aot}                                                      & -               & -              & 83.7           & 8.3                                             \\
Share LSTT layers                                                    & -               & -              & 82.8           & \textbf{5.7}                                    \\ \hline
\end{tabular}
\end{subtable}
\begin{subtable}{.35\textwidth}
\center
\caption{Layer-wise ID-based attention}\label{tab:lstt_block_v2}\vspace{-1mm}
\setlength{\tabcolsep}{5pt}

\begin{tabular}{|c|c|ccc|c|}
\hline
\rowcolor[HTML]{FCFCFC} 
\cellcolor[HTML]{FCFCFC}                               & \cellcolor[HTML]{FCFCFC}                                 & \multicolumn{3}{c|}{\cellcolor[HTML]{FCFCFC}$n_l'$}                                                             & \cellcolor[HTML]{FCFCFC}                        \\ \cline{3-5}
\rowcolor[HTML]{FCFCFC} 
\multirow{-2}{*}{\cellcolor[HTML]{FCFCFC}$W^{id}_{l}$} & \multirow{-2}{*}{\cellcolor[HTML]{FCFCFC}$W^{gate}_{l}$} & \multicolumn{1}{c|}{\cellcolor[HTML]{FCFCFC}1} & \multicolumn{1}{c|}{\cellcolor[HTML]{FCFCFC}2} & 3             & \multirow{-2}{*}{\cellcolor[HTML]{FCFCFC}Param} \\ \hline
\checkmark                                             & \checkmark                                               & \textbf{80.6}                                  & \textbf{83.7}                                  & \textbf{84.4} & \underline{8.8}                                       \\ \hline
                                                       & \checkmark                                               & 80.0                                           & 83.2                                           & 83.7          & \textbf{8.7}                                    \\
\checkmark                                             &                                                          & \underline{80.3}                                     & \underline{83.5}                                     & \underline{84.1}    & \underline{8.8}                                       \\
                                                       &                                                          & 79.8                                           & 82.6                                           & 83.4          & \textbf{8.7}                                    \\ \hline
\end{tabular}

\end{subtable}
\vspace{-4.0mm}
\end{table*}

\noindent\textbf{Layer-wise ID-based Attention:} We further validate the necessity of identification weight $W^{id}_{l}$ and gating weight $W^{gate}_{l}$ in Table~\ref{tab:lstt_block_v2}. In detail, including $W^{id}_{l}$ or $W^{gate}_{l}$ in S-LSTT undoubtedly warrants performance improvements but only marginally increases the network parameters. Particularly, the improvements obtained from $W^{id}_{l}$ (80.3/83.5/84.1\% \vs 79.8/82.6/83.4\%) are more significant than those obtained from $W^{gate}_{l}$ (80.0/83.3/83.7\% \vs 79.8/82.6/83.4\%). In other words, incorporating layer-wise ID embeddings in different blocks is more critical in enhancing the representative ability of S-LSTT. 


\vspace{-2mm}

\section{Conclusion}

This paper engages in a pioneering endeavor to emphasize the importance of multi-object modeling and flexible deployment in VOS. We propose two innovative and effective approaches for video object segmentation using scalable long short-term transformers to associate identification embeddings of multiple objects. With the identification mechanism, AOT can execute more effective multi-object training and inference, while efficiently as their single-object counterparts. With layer-wise ID-based attention and scalable supervision, AOST further overcomes ID embeddings’ representation limitations and enables online architecture scalability in VOS for the first time. 
Since there is no standard VOS benchmark involving scenarios with dense objects, we carefully present a challenging multi-object benchmark, VOSW.
Remarkably, the proposed AOST and AOT approaches achieve superior performance on VOSW and five commonly-used benchmarks, demonstrating the importance of multi-object modeling and the effectiveness of AOST's scalable architecture. We hope our approaches and benchmark foster future studies on multi-object VOS and related tasks (\eg, video instance segmentation and interactive VOS). Moreover, our AOST \& AOT variants can serve as firm baselines for such investigations.

\ifCLASSOPTIONcaptionsoff
  \newpage
\fi



\small{
\bibliographystyle{IEEEtran}
\bibliography{reference}

\begin{thebibliography}{100}
\providecommand{\url}[1]{#1}
\csname url@samestyle\endcsname
\providecommand{\newblock}{\relax}
\providecommand{\bibinfo}[2]{#2}
\providecommand{\BIBentrySTDinterwordspacing}{\spaceskip=0pt\relax}
\providecommand{\BIBentryALTinterwordstretchfactor}{4}
\providecommand{\BIBentryALTinterwordspacing}{\spaceskip=\fontdimen2\font plus
\BIBentryALTinterwordstretchfactor\fontdimen3\font minus \fontdimen4\font\relax}
\providecommand{\BIBforeignlanguage}[2]{{%
\expandafter\ifx\csname l@#1\endcsname\relax
\typeout{** WARNING: IEEEtran.bst: No hyphenation pattern has been}%
\typeout{** loaded for the language `#1'. Using the pattern for}%
\typeout{** the default language instead.}%
\else
\language=\csname l@#1\endcsname
\fi
#2}}
\providecommand{\BIBdecl}{\relax}
\BIBdecl

\bibitem{wang2021survey}
T.~Zhou, F.~Porikli, D.~J. Crandall, L.~Van~Gool, and W.~Wang, ``A survey on deep learning technique for video segmentation,'' \emph{TPAMI}, 2022.

\bibitem{lecun2015deep}
Y.~LeCun, Y.~Bengio, and G.~Hinton, ``Deep learning,'' \emph{nature}, vol. 521, no. 7553, pp. 436--444, 2015.

\bibitem{spacetime}
S.~W. Oh, J.-Y. Lee, N.~Xu, and S.~J. Kim, ``Video object segmentation using space-time memory networks,'' in \emph{ICCV}, 2019.

\bibitem{KMN}
H.~Seong, J.~Hyun, and E.~Kim, ``Kernelized memory network for video object segmentation,'' in \emph{ECCV}, 2020.

\bibitem{EGMN}
X.~Lu, W.~Wang, M.~Danelljan, T.~Zhou, J.~Shen, and L.~Van~Gool, ``Video object segmentation with episodic graph memory networks,'' in \emph{ECCV}, 2020.

\bibitem{cheng2021stcn}
H.~K. Cheng, Y.-W. Tai, and C.-K. Tang, ``Rethinking space-time networks with improved memory coverage for efficient video object segmentation,'' in \emph{NeurIPS}, 2021.

\bibitem{nonlocal}
X.~Wang, R.~Girshick, A.~Gupta, and K.~He, ``Non-local neural networks,'' in \emph{CVPR}, 2018, pp. 7794--7803.

\bibitem{cfbi}
Z.~Yang, Y.~Wei, and Y.~Yang, ``Collaborative video object segmentation by foreground-background integration,'' in \emph{ECCV}, 2020.

\bibitem{cfbip}
Z.~Yang, Y.~Wei, and Y.~Yang\;\!\!, ``Collaborative video object segmentation by multi-scale foreground-background integration,'' \emph{TPAMI}, vol.~44, no.~9, pp. 4701--4712, 2021.

\bibitem{rpcm}
X.~Xu, J.~Wang, X.~Li, and Y.~Lu, ``Reliable propagation-correction modulation for video object segmentation,'' in \emph{AAAI}, 2022.

\bibitem{cho2022tackling}
S.~Cho, H.~Lee, M.~Lee, C.~Park, S.~Jang, M.~Kim, and S.~Lee, ``Tackling background distraction in video object segmentation,'' in \emph{ECCV}, 2022, pp. 446--462.

\bibitem{feelvos}
P.~Voigtlaender, Y.~Chai, F.~Schroff, H.~Adam, B.~Leibe, and L.-C. Chen, ``Feelvos: Fast end-to-end embedding learning for video object segmentation,'' in \emph{CVPR}, 2019, pp. 9481--9490.

\bibitem{miles2023mobilevos}
R.~Miles, M.~K. Yucel, B.~Manganelli, and A.~Sa{\`a}-Garriga, ``Mobilevos: Real-time video object segmentation contrastive learning meets knowledge distillation,'' in \emph{CVPR}, 2023, pp. 10\,480--10\,490.

\bibitem{realtimevos2}
Y.~Li, Z.~Shen, and Y.~Shan, ``Fast video object segmentation using the global context module,'' in \emph{ECCV}, 2020, pp. 735--750.

\bibitem{realtimevos1}
X.~Chen, Z.~Li, Y.~Yuan, G.~Yu, J.~Shen, and D.~Qi, ``State-aware tracker for real-time video object segmentation,'' in \emph{CVPR}, 2020, pp. 9384--9393.

\bibitem{NEURIPS2020_liangVOS}
Y.~Liang, X.~Li, N.~Jafari, and J.~Chen, ``Video object segmentation with adaptive feature bank and uncertain-region refinement,'' in \emph{NeurIPS}, vol.~33, 2020, pp. 3430--3441.

\bibitem{aot}
Z.~Yang, Y.~Wei, and Y.~Yang, ``Associating objects with transformers for video object segmentation,'' in \emph{NeurIPS}, 2021.

\bibitem{miao2022large}
J.~Miao, X.~Wang, Y.~Wu, W.~Li, X.~Zhang, Y.~Wei, and Y.~Yang, ``Large-scale video panoptic segmentation in the wild: A benchmark,'' in \emph{CVPR}, 2022, pp. 21\,033--21\,043.

\bibitem{youtubevos}
N.~Xu, L.~Yang, Y.~Fan, D.~Yue, Y.~Liang, J.~Yang, and T.~Huang, ``Youtube-vos: A large-scale video object segmentation benchmark,'' \emph{arXiv preprint arXiv:1809.03327}, 2018.

\bibitem{davis2017}
J.~Pont-Tuset, F.~Perazzi, S.~Caelles, P.~Arbel{\'a}ez, A.~Sorkine-Hornung, and L.~Van~Gool, ``The 2017 davis challenge on video object segmentation,'' \emph{arXiv preprint arXiv:1704.00675}, 2017.

\bibitem{davis2016}
F.~Perazzi, J.~Pont-Tuset, B.~McWilliams, L.~Van~Gool, M.~Gross, and A.~Sorkine-Hornung, ``A benchmark dataset and evaluation methodology for video object segmentation,'' in \emph{CVPR}, 2016, pp. 724--732.

\bibitem{sandler2018mobilenetv2}
M.~Sandler, A.~Howard, M.~Zhu, A.~Zhmoginov, and L.-C. Chen, ``Mobilenetv2: Inverted residuals and linear bottlenecks,'' in \emph{CVPR}, 2018, pp. 4510--4520.

\bibitem{swin}
Z.~Liu, Y.~Lin, Y.~Cao, H.~Hu, Y.~Wei, Z.~Zhang, S.~Lin, and B.~Guo, ``Swin transformer: Hierarchical vision transformer using shifted windows,'' in \emph{ICCV}, 2021.

\bibitem{aot_workshop}
Z.~Yang, J.~Zhang, W.~Wang, W.~Han, Y.~Yu, Y.~Li, J.~Wang, Y.~Wei, Y.~Sun, and Y.~Yang, ``Towards multi-object association from foreground-background integration,'' in \emph{CVPR Workshops}, 2021.

\bibitem{zhu2022instance}
F.~Zhu, Z.~Yang, X.~Yu, Y.~Yang, and Y.~Wei, ``Instance as identity: A generic online paradigm for video instance segmentation,'' in \emph{ECCV}, 2022, pp. 524--540.

\bibitem{cheng2023segment}
Y.~Cheng, L.~Li, Y.~Xu, X.~Li, Z.~Yang, W.~Wang, and Y.~Yang, ``Segment and track anything,'' \emph{arXiv preprint arXiv:2305.06558}, 2023.

\bibitem{yang2022decoupling}
Z.~Yang and Y.~Yang, ``Decoupling features in hierarchical propagation for video object segmentation,'' in \emph{NeurIPS}, 2022.

\bibitem{xu2023integrating}
Y.~Xu, Z.~Yang, and Y.~Yang, ``Integrating boxes and masks: A multi-object framework for unified visual tracking and segmentation,'' in \emph{ICCV}, 2023, pp. 9738--9751.

\bibitem{yang2024doraemongpt}
Z.~Yang, G.~Chen, X.~Li, W.~Wang, and Y.~Yang, ``Doraemongpt: Toward understanding dynamic scenes with large language models,'' 2024.

\bibitem{li2023catr}
K.~Li, Z.~Yang, L.~Chen, Y.~Yang, and J.~Xiao, ``Catr: Combinatorial-dependence audio-queried transformer for audio-visual video segmentation,'' in \emph{Proceedings of the 31st ACM International Conference on Multimedia}, 2023, pp. 1485--1494.

\bibitem{xu2023video}
Y.~Xu, Z.~Yang, and Y.~Yang\;\!\!, ``Video object segmentation in panoptic wild scenes,'' in \emph{IJCAI}, 2023.

\bibitem{yu2022batman}
Y.~Yu, J.~Yuan, G.~Mittal, L.~Fuxin, and M.~Chen, ``Batman: Bilateral attention transformer in motion-appearance neighboring space for video object segmentation,'' in \emph{ECCV}, 2022, pp. 612--629.

\bibitem{xu2023mbptrack}
T.-X. Xu, Y.-C. Guo, Y.-K. Lai, and S.-H. Zhang, ``Mbptrack: Improving 3d point cloud tracking with memory networks and box priors,'' \emph{arXiv preprint arXiv:2303.05071}, 2023.

\bibitem{tokmakov2023breaking}
P.~Tokmakov, J.~Li, and A.~Gaidon, ``Breaking the" object" in video object segmentation,'' in \emph{CVPR}, 2023, pp. 22\,836--22\,845.

\bibitem{mayer2022beyond}
C.~Mayer, M.~Danelljan, M.-H. Yang, V.~Ferrari, L.~Van~Gool, and A.~Kuznetsova, ``Beyond sot: It's time to track multiple generic objects at once,'' in \emph{WACV}, 2024.

\bibitem{chen2015video}
L.~Chen, J.~Shen, W.~Wang, and B.~Ni, ``Video object segmentation via dense trajectories,'' \emph{TMM}, vol.~17, no.~12, pp. 2225--2234, 2015.

\bibitem{liang2023local}
C.~Liang, W.~Wang, T.~Zhou, J.~Miao, Y.~Luo, and Y.~Yang, ``Local-global context aware transformer for language-guided video segmentation,'' \emph{TPAMI}, 2023.

\bibitem{tradition1}
V.~Badrinarayanan, F.~Galasso, and R.~Cipolla, ``Label propagation in video sequences,'' in \emph{CVPR}, 2010, pp. 3265--3272.

\bibitem{tradition3}
S.~Vijayanarasimhan and K.~Grauman, ``Active frame selection for label propagation in videos,'' in \emph{ECCV}, 2012, pp. 496--509.

\bibitem{tradition2}
S.~Avinash~Ramakanth and R.~Venkatesh~Babu, ``Seamseg: Video object segmentation using patch seams,'' in \emph{CVPR}, 2014, pp. 376--383.

\bibitem{osvos}
S.~Caelles, K.-K. Maninis, J.~Pont-Tuset, L.~Leal-Taix{\'e}, D.~Cremers, and L.~Van~Gool, ``One-shot video object segmentation,'' in \emph{CVPR}, 2017, pp. 221--230.

\bibitem{xiao2018monet}
H.~Xiao, J.~Feng, G.~Lin, Y.~Liu, and M.~Zhang, ``Monet: Deep motion exploitation for video object segmentation,'' in \emph{CVPR}, 2018, pp. 1140--1148.

\bibitem{onavos}
P.~Voigtlaender and B.~Leibe, ``Online adaptation of convolutional neural networks for video object segmentation,'' in \emph{BMVC}, 2017.

\bibitem{masktrack}
F.~Perazzi, A.~Khoreva, R.~Benenson, B.~Schiele, and A.~Sorkine-Hornung, ``Learning video object segmentation from static images,'' in \emph{CVPR}, 2017, pp. 2663--2672.

\bibitem{premvos}
J.~Luiten, P.~Voigtlaender, and B.~Leibe, ``Premvos: Proposal-generation, refinement and merging for video object segmentation,'' in \emph{ACCV}, 2018, pp. 565--580.

\bibitem{osmn}
L.~Yang, Y.~Wang, X.~Xiong, J.~Yang, and A.~K. Katsaggelos, ``Efficient video object segmentation via network modulation,'' in \emph{CVPR}, 2018, pp. 6499--6507.

\bibitem{pml}
Y.~Chen, J.~Pont-Tuset, A.~Montes, and L.~Van~Gool, ``Blazingly fast video object segmentation with pixel-wise metric learning,'' in \emph{CVPR}, 2018, pp. 1189--1198.

\bibitem{videomatch}
Y.-T. Hu, J.-B. Huang, and A.~G. Schwing, ``Videomatch: Matching based video object segmentation,'' in \emph{ECCV}, 2018, pp. 54--70.

\bibitem{yang2019going}
Z.~Yang, P.~Li, Q.~Feng, Y.~Wei, and Y.~Yang, ``Going deeper into embedding learning for video object segmentation,'' in \emph{ICCV Workshops}, 2019, pp. 0--0.

\bibitem{yang2020cfbi+}
Z.~Yang, Y.~Ding, Y.~Wei, and Y.~Yang, ``Cfbi+: Collaborative video object segmentation by multi-scale foreground-background integration,'' in \emph{CVPR Workshops}, vol.~1, no.~2, 2020, p.~3.

\bibitem{rgmp}
S.~Wug~Oh, J.-Y. Lee, K.~Sunkavalli, and S.~Joo~Kim, ``Fast video object segmentation by reference-guided mask propagation,'' in \emph{CVPR}, 2018, pp. 7376--7385.

\bibitem{LWLVOS}
G.~Bhat, F.~J. Lawin, M.~Danelljan, A.~Robinson, M.~Felsberg, L.~Van~Gool, and R.~Timofte, ``Learning what to learn for video object segmentation,'' in \emph{ECCV}, 2020.

\bibitem{mao2021joint}
Y.~Mao, N.~Wang, W.~Zhou, and H.~Li, ``Joint inductive and transductive learning for video object segmentation,'' in \emph{ICCV}, 2021, pp. 9670--9679.

\bibitem{att}
D.~Bahdanau, K.~Cho, and Y.~Bengio, ``Neural machine translation by jointly learning to align and translate,'' in \emph{ICLR}, 2015.

\bibitem{transformer}
A.~Vaswani, N.~Shazeer, N.~Parmar, J.~Uszkoreit, L.~Jones, A.~N. Gomez, L.~Kaiser, and I.~Polosukhin, ``Attention is all you need,'' in \emph{NeurIPS}, 2017.

\bibitem{seong2022video}
H.~Seong, J.~Hyun, and E.~Kim, ``Video object segmentation using kernelized memory network with multiple kernels,'' \emph{TPAMI}, 2022.

\bibitem{cheng2022xmem}
H.~K. Cheng and A.~G. Schwing, ``Xmem: Long-term video object segmentation with an atkinson-shiffrin memory model,'' in \emph{ECCV}, 2022, pp. 640--658.

\bibitem{zhang2023boosting}
Y.~Zhang, L.~Li, W.~Wang, R.~Xie, L.~Song, and W.~Zhang, ``Boosting video object segmentation via space-time correspondence learning,'' in \emph{CVPR}, 2023, pp. 2246--2256.

\bibitem{rde_cvpr22}
M.~Li, L.~Hu, Z.~Xiong, B.~Zhang, P.~Pan, and D.~Liu, ``Recurrent dynamic embedding for video object segmentation,'' in \emph{CVPR}, 2022, pp. 1332--1341.

\bibitem{wang2023look}
J.~Wang, D.~Chen, Z.~Wu, C.~Luo, C.~Tang, X.~Dai, Y.~Zhao, Y.~Xie, L.~Yuan, and Y.-G. Jiang, ``Look before you match: Instance understanding matters in video object segmentation,'' in \emph{CVPR}, 2023, pp. 2268--2278.

\bibitem{wang2018semi}
W.~Wang, J.~Shen, F.~Porikli, and R.~Yang, ``Semi-supervised video object segmentation with super-trajectories,'' \emph{TPAMI}, vol.~41, no.~4, pp. 985--998, 2018.

\bibitem{teerapittayanon2016branchynet}
S.~Teerapittayanon, B.~McDanel, and H.-T. Kung, ``Branchynet: Fast inference via early exiting from deep neural networks,'' in \emph{ICPR}.\hskip 1em plus 0.5em minus 0.4em\relax IEEE, 2016, pp. 2464--2469.

\bibitem{bolukbasi2017adaptive}
T.~Bolukbasi, J.~Wang, O.~Dekel, and V.~Saligrama, ``Adaptive neural networks for efficient inference,'' in \emph{ICML}.\hskip 1em plus 0.5em minus 0.4em\relax PMLR, 2017, pp. 527--536.

\bibitem{yu2018slimmable}
J.~Yu, L.~Yang, N.~Xu, J.~Yang, and T.~Huang, ``Slimmable neural networks,'' in \emph{ICLR}, 2018.

\bibitem{ln}
J.~L. Ba, J.~R. Kiros, and G.~E. Hinton, ``Layer normalization,'' in \emph{NeurIPS Workshops}, 2016.

\bibitem{devlin2018bert}
J.~Devlin, M.-W. Chang, K.~Lee, and K.~Toutanova, ``Bert: Pre-training of deep bidirectional transformers for language understanding,'' in \emph{NAACL}, 2019, pp. 4171–--4186.

\bibitem{radford2019language}
A.~Radford, J.~Wu, R.~Child, D.~Luan, D.~Amodei, and I.~Sutskever, ``Language models are unsupervised multitask learners,'' \emph{OpenAI blog}, vol.~1, no.~8, p.~9, 2019.

\bibitem{synnaeve2019end}
G.~Synnaeve, Q.~Xu, J.~Kahn, T.~Likhomanenko, E.~Grave, V.~Pratap, A.~Sriram, V.~Liptchinsky, and R.~Collobert, ``End-to-end asr: from supervised to semi-supervised learning with modern architectures,'' in \emph{ICML Workshops}, 2020.

\bibitem{vit}
A.~Dosovitskiy, L.~Beyer, A.~Kolesnikov, D.~Weissenborn, X.~Zhai, T.~Unterthiner, M.~Dehghani, M.~Minderer, G.~Heigold, S.~Gelly \emph{et~al.}, ``An image is worth 16x16 words: Transformers for image recognition at scale,'' in \emph{ICLR}, 2021.

\bibitem{vaswani2021scaling}
A.~Vaswani, P.~Ramachandran, A.~Srinivas, N.~Parmar, B.~Hechtman, and J.~Shlens, ``Scaling local self-attention for parameter efficient visual backbones,'' in \emph{CVPR}, 2021, pp. 12\,894--12\,904.

\bibitem{detr}
N.~Carion, F.~Massa, G.~Synnaeve, N.~Usunier, A.~Kirillov, and S.~Zagoruyko, ``End-to-end object detection with transformers,'' in \emph{ECCV}, 2020, pp. 213--229.

\bibitem{vistr}
Y.~Wang, Z.~Xu, X.~Wang, C.~Shen, B.~Cheng, H.~Shen, and H.~Xia, ``End-to-end video instance segmentation with transformers,'' in \emph{CVPR}, 2021, pp. 8741--8750.

\bibitem{parmar2018image}
N.~Parmar, A.~Vaswani, J.~Uszkoreit, L.~Kaiser, N.~Shazeer, A.~Ku, and D.~Tran, ``Image transformer,'' in \emph{ICCV}, 2018, pp. 4055--4064.

\bibitem{arnab2021vivit}
A.~Arnab, M.~Dehghani, G.~Heigold, C.~Sun, M.~Lu{\v{c}}i{\'c}, and C.~Schmid, ``Vivit: A video vision transformer,'' in \emph{ICCV}, 2021, pp. 6836--6846.

\bibitem{zhu2021temporal}
L.~Zhu, H.~Fan, Y.~Luo, M.~Xu, and Y.~Yang, ``Temporal cross-layer correlation mining for action recognition,'' \emph{TMM}, vol.~24, pp. 668--676, 2021.

\bibitem{lu2023show}
Y.~Lu, F.~Ni, H.~Wang, X.~Guo, L.~Zhu, Z.~Yang, R.~Song, L.~Cheng, and Y.~Yang, ``Show me a video: A large-scale narrated video dataset for coherent story illustration,'' \emph{TMM}, 2023.

\bibitem{wang2022align}
X.~Wang, L.~Zhu, Z.~Zheng, M.~Xu, and Y.~Yang, ``Align and tell: Boosting text-video retrieval with local alignment and fine-grained supervision,'' 2022.

\bibitem{bertasius2021space}
G.~Bertasius, H.~Wang, and L.~Torresani, ``Is space-time attention all you need for video understanding?'' in \emph{ICML}, vol.~2, no.~3, 2021, p.~4.

\bibitem{fan2021multiscale}
H.~Fan, B.~Xiong, K.~Mangalam, Y.~Li, Z.~Yan, J.~Malik, and C.~Feichtenhofer, ``Multiscale vision transformers,'' in \emph{ICCV}, 2021, pp. 6824--6835.

\bibitem{liu2022video}
Z.~Liu, J.~Ning, Y.~Cao, Y.~Wei, Z.~Zhang, S.~Lin, and H.~Hu, ``Video swin transformer,'' in \emph{CVPR}, 2022, pp. 3202--3211.

\bibitem{wang2021transformer}
N.~Wang, W.~Zhou, J.~Wang, and H.~Li, ``Transformer meets tracker: Exploiting temporal context for robust visual tracking,'' in \emph{CVPR}, 2021, pp. 1571--1580.

\bibitem{chen2021transformer}
X.~Chen, B.~Yan, J.~Zhu, D.~Wang, X.~Yang, and H.~Lu, ``Transformer tracking,'' in \emph{CVPR}, 2021, pp. 8126--8135.

\bibitem{yan2023universal}
B.~Yan, Y.~Jiang, J.~Wu, D.~Wang, P.~Luo, Z.~Yuan, and H.~Lu, ``Universal instance perception as object discovery and retrieval,'' \emph{arXiv preprint arXiv:2303.06674}, 2023.

\bibitem{yan2022towards}
B.~Yan, Y.~Jiang, P.~Sun, D.~Wang, Z.~Yuan, P.~Luo, and H.~Lu, ``Towards grand unification of object tracking,'' in \emph{ECCV}, 2022, pp. 733--751.

\bibitem{meinhardt2022trackformer}
T.~Meinhardt, A.~Kirillov, L.~Leal-Taixe, and C.~Feichtenhofer, ``Trackformer: Multi-object tracking with transformers,'' in \emph{CVPR}, 2022, pp. 8844--8854.

\bibitem{chu2023transmot}
P.~Chu, J.~Wang, Q.~You, H.~Ling, and Z.~Liu, ``Transmot: Spatial-temporal graph transformer for multiple object tracking,'' in \emph{WACV}, 2023, pp. 4870--4880.

\bibitem{lin2019agss}
H.~Lin, X.~Qi, and J.~Jia, ``Agss-vos: Attention guided single-shot video object segmentation,'' in \emph{ICCV}, 2019, pp. 3949--3957.

\bibitem{liu2022global}
Y.~Liu, R.~Yu, J.~Wang, X.~Zhao, Y.~Wang, Y.~Tang, and Y.~Yang, ``Global spectral filter memory network for video object segmentation,'' in \emph{ECCV}, 2022, pp. 648--665.

\bibitem{liu2022learning}
Y.~Liu, R.~Yu, F.~Yin, X.~Zhao, W.~Zhao, W.~Xia, and Y.~Yang, ``Learning quality-aware dynamic memory for video object segmentation,'' in \emph{ECCV}, 2022, pp. 468--486.

\bibitem{gelu}
D.~Hendrycks and K.~Gimpel, ``Gaussian error linear units (gelus),'' \emph{arXiv preprint arXiv:1606.08415}, 2016.

\bibitem{nowozin2014optimal}
S.~Nowozin, ``Optimal decisions from probabilistic models: the intersection-over-union case,'' in \emph{CVPR}, 2014, pp. 548--555.

\bibitem{gct}
Z.~Yang, L.~Zhu, Y.~Wu, and Y.~Yang, ``Gated channel transformation for visual recognition,'' in \emph{CVPR}, 2020.

\bibitem{fpn}
T.-Y. Lin, P.~Doll{\'a}r, R.~Girshick, K.~He, B.~Hariharan, and S.~Belongie, ``Feature pyramid networks for object detection,'' in \emph{CVPR}, 2017, pp. 2117--2125.

\bibitem{gn}
Y.~Wu and K.~He, ``Group normalization,'' in \emph{ECCV}, 2018, pp. 3--19.

\bibitem{resnet}
K.~He, X.~Zhang, S.~Ren, and J.~Sun, ``Deep residual learning for image recognition,'' in \emph{CVPR}, 2016.

\bibitem{voc}
M.~Everingham, L.~Van~Gool, C.~K. Williams, J.~Winn, and A.~Zisserman, ``The pascal visual object classes (voc) challenge,'' \emph{IJCV}, vol.~88, no.~2, pp. 303--338, 2010.

\bibitem{coco}
T.-Y. Lin, M.~Maire, S.~Belongie, J.~Hays, P.~Perona, D.~Ramanan, P.~Doll{\'a}r, and C.~L. Zitnick, ``Microsoft coco: Common objects in context,'' in \emph{ECCV}, 2014, pp. 740--755.

\bibitem{cheng2014global}
M.-M. Cheng, N.~J. Mitra, X.~Huang, P.~H. Torr, and S.-M. Hu, ``Global contrast based salient region detection,'' \emph{TPAMI}, vol.~37, no.~3, pp. 569--582, 2014.

\bibitem{shi2015hierarchical}
J.~Shi, Q.~Yan, L.~Xu, and J.~Jia, ``Hierarchical image saliency detection on extended cssd,'' \emph{TPAMI}, vol.~38, no.~4, pp. 717--729, 2015.

\bibitem{semantic}
B.~Hariharan, P.~Arbel{\'a}ez, L.~Bourdev, S.~Maji, and J.~Malik, ``Semantic contours from inverse detectors,'' in \emph{ICCV}, 2011, pp. 991--998.

\bibitem{adamw}
I.~Loshchilov and F.~Hutter, ``Decoupled weight decay regularization,'' in \emph{ICLR}, 2019.

\bibitem{bn}
S.~Ioffe and C.~Szegedy, ``Batch normalization: Accelerating deep network training by reducing internal covariate shift,'' in \emph{ICML}, 2015.

\bibitem{polyak1992acceleration}
B.~T. Polyak and A.~B. Juditsky, ``Acceleration of stochastic approximation by averaging,'' \emph{SIAM journal on control and optimization}, vol.~30, no.~4, pp. 838--855, 1992.

\bibitem{huang2016deep}
G.~Huang, Y.~Sun, Z.~Liu, D.~Sedra, and K.~Q. Weinberger, ``Deep networks with stochastic depth,'' in \emph{ECCV}, 2016, pp. 646--661.

\bibitem{wang2021unidentified}
W.~Wang, M.~Feiszli, H.~Wang, and D.~Tran, ``Unidentified video objects: A benchmark for dense, open-world segmentation,'' in \emph{ICCV}, 2021, pp. 10\,776--10\,785.

\bibitem{pcvos_cvpr22}
K.~Park, S.~Woo, S.~W. Oh, I.~S. Kweon, and J.-Y. Lee, ``Per-clip video object segmentation,'' in \emph{CVPR}, 2022, pp. 1352--1361.

\bibitem{sstvos}
B.~Duke, A.~Ahmed, C.~Wolf, P.~Aarabi, and G.~W. Taylor, ``Sstvos: Sparse spatiotemporal transformers for video object segmentation,'' in \emph{CVPR}, 2021.

\bibitem{hmmn}
H.~Seong, S.~W. Oh, J.-Y. Lee, S.~Lee, S.~Lee, and E.~Kim, ``Hierarchical memory matching network for video object segmentation,'' in \emph{ICCV}, 2021, pp. 12\,889--12\,898.

\bibitem{llb_aaai23}
M.~Lan, J.~Zhang, L.~Zhang, and D.~Tao, ``Learning to learn better for video object segmentation,'' in \emph{AAAI}, 2023.

\end{thebibliography}
}
%


%







\newpage

\end{document}